\documentclass[10pt,twocolumn,letterpaper]{article}
\usepackage[pagenumbers]{cvpr}
\usepackage{amsfonts}
\usepackage{algorithm}
\usepackage{algorithmic}
\usepackage{array}
\usepackage{multirow}
\usepackage{textcomp}
\usepackage{CJKutf8}
\usepackage{verbatim}
\definecolor{cvprblue}{rgb}{0.21,0.49,0.74}
\usepackage[pagebackref,breaklinks,colorlinks,allcolors=cvprblue]{hyperref}
\hypersetup{
  pdfauthor={Xiaojun Bi, Xiaoyuan Ma, Yiwen Sun, Tianren Huang, Chaoran Liu, Bokai Huang, Hao Yang, Baichuan Mo}
}
\def\confName{CVPR}
\def\confYear{2027}

\begin{document}

\title{Retrieval-Augmented Large Language Model Decision-Making for\\Autonomous Driving Guided by Chinese Philosophical Wisdom}

\author{%
\normalfont
\begin{minipage}[t]{\dimexpr\textwidth-2\tabcolsep\relax}
\centering
\textbf{Xiaojun Bi}\textsuperscript{1}\quad
\textbf{Xiaoyuan Ma}\textsuperscript{1}\quad
\textbf{Yiwen Sun}\textsuperscript{2,3,*}\quad
\textbf{Tianren Huang}\textsuperscript{4}\\[3pt]
\textbf{Chaoran Liu}\textsuperscript{5}\quad
\textbf{Bokai Huang}\textsuperscript{6}\quad
\textbf{Hao Yang}\textsuperscript{8}\quad
\textbf{Baichuan Mo}\textsuperscript{7}\\[7pt]
\small
\textsuperscript{1}Key Laboratory of Ethnic Language Intelligent Analysis and Security Governance of MOE,\\
and School of Information and Engineering, Minzu University of China, Beijing 100081, China\\[2pt]
\textsuperscript{2}School of Intelligence Science and Technology, Peking University, Beijing 100871, China\\[2pt]
\textsuperscript{3}State Key Laboratory of General Artificial Intelligence, BIGAI, Beijing 100080, China\\[2pt]
\textsuperscript{4}School of Electronics Engineering and Computer Science,\\
Peking University, Beijing 100871, China\\[2pt]
\textsuperscript{5}School of Earth and Space Sciences, Peking University, Beijing 100871, China\\[2pt]
\textsuperscript{6}Yuanpei College, Peking University, Beijing 100871, China\\[2pt]
\textsuperscript{7}Department of Civil Engineering, Tsinghua University, Beijing 100084, China\\[2pt]
\textsuperscript{8}Institute for Artificial Intelligence, Peking University, Beijing 100871, China\\[5pt]
\textsuperscript{*}Corresponding author: Yiwen Sun (\href{mailto:sunyiwen@pku.edu.cn}{sunyiwen@pku.edu.cn})
\end{minipage}%
}

\maketitle

\begin{abstract}
Autonomous driving decision systems must balance safety, efficiency, and social norms in complex traffic interactions. Philosophical and ethical considerations have received limited attention in existing autonomous driving decision-making approaches based on numerical optimization, sequence prediction, and large language models (LLMs). We propose Chinese Philosophical Wisdom-Guided Driving (CPW-Drive), a closed-loop retrieval-augmented generation (RAG) framework that incorporates value guidance derived from Chinese philosophy into autonomous driving decision-making. Using Chinese Confucian thought as its knowledge source, CPW-Drive consolidates LLM-extracted keywords from relevant classical texts into driving-relevant value principles through manual screening and validation. It then contextualizes these principles through scenario-specific cases to form retrievable and reusable value guidance. We further propose Physics-aware Spatial Similarity Retrieval (PSSR), which compares vehicle layouts and velocity-extrapolated states to retrieve physically relevant historical cases. On Highway-env's multilane highway-driving task, CPW-Drive achieves success rates of 93.0\%, 86.0\%, and 72.0\% across three traffic configurations. These results outperform the strongest baseline by 8.0, 22.5, and 25.0 percentage points, respectively. Across all configurations, CPW-Drive achieves the highest collision-free step count and maintains a low lane-change frequency. The results suggest that structured value guidance can improve simulated closed-loop safety and stability, with an efficiency trade-off. The source code will be made publicly available upon publication.
\end{abstract}


\section{Introduction}

Recent advances in autonomous driving have improved perception, planning, and control. However, data-driven approaches, including reinforcement learning (RL) and imitation learning (IL), still face interpretability and generalization challenges in open-ended, long-tail scenarios~\citep{kiran2021deep,zhu2021survey,tian2024tokenize}. Moreover, numerical optimization and prediction objectives do not explicitly encode every social norm that can influence human driving. Interactive traffic therefore requires trade-offs among safety, efficiency, and traffic etiquette.

Knowledge-driven autonomous driving has consequently attracted increasing attention. Recent studies have explored large language models (LLMs) for driving decisions because they can combine language-based reasoning, external knowledge, and scene descriptions~\citep{mao2023language,wen2024dilu}. Most LLM-based driving methods rely on task instructions, commonsense knowledge, or historical experience~\citep{mao2023gpt,sha2023languagempc,wen2024dilu}. More recent systems retrieve traffic regulations or learn value preferences~\citep{cai2026driving,xia2026knowval}. However, the literature has not established how culturally grounded philosophical principles can be represented as reusable, scenario-level driving examples. This gap limits the systematic application of abstract value principles to concrete traffic interactions.

% To address this gap, we propose CPW-Drive, an LLM-based framework that integrates Chinese philosophical wisdom into autonomous driving decisions. The framework transforms abstract values into structured, reusable driving knowledge for concrete traffic scenarios. We select Confucian thought as the knowledge source and operationalize themes of order, responsibility, restraint, and relational coordination. Semantic Anchors and a Philosophical Driving Knowledge Base connect these themes to reasoning processes and driving actions. Physics-aware Spatial Similarity Retrieval (PSSR) then supplies relevant historical cases for each decision. Figure~\ref{fig:intro_concept} illustrates the overall concept.

\begin{figure}[t]
\centering
\includegraphics[width=\linewidth]{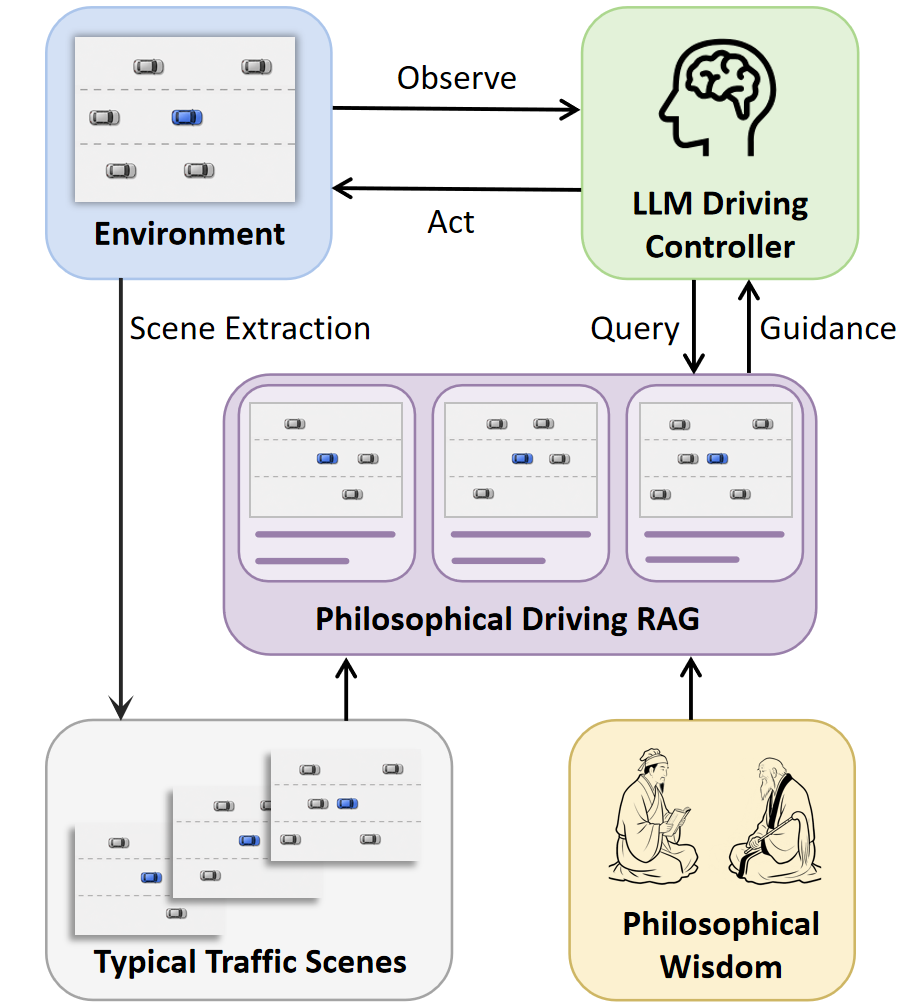}
\caption{Conceptual illustration of CPW-Drive. Driving-relevant value principles derived from Chinese philosophical wisdom provide retrievable guidance for closed-loop LLM-based driving decisions.}
\label{fig:intro_concept}
\end{figure}
To address this gap, we propose Chinese Philosophical Wisdom-Guided Driving (CPW-Drive), an LLM-based framework that integrates Chinese philosophical wisdom into autonomous driving decision-making. The framework transforms abstract values into structured driving knowledge for concrete traffic scenarios. We select Confucian thought as the knowledge source, extract high-frequency keywords from relevant classical texts, and consolidate them into ten driving-relevant value principles through manual screening and conceptual integration. These value principles are then combined with specific driving scenarios to form retrievable and reusable value-guided driving cases, while Physics-aware Spatial Similarity Retrieval (PSSR) retrieves relevant historical cases for the current decision. Figure~\ref{fig:intro_concept} illustrates the overall concept.

The main contributions of this paper are summarized as follows:

\begin{itemize}
\item To the best of our knowledge, CPW-Drive is the first LLM-based autonomous driving framework to incorporate driving-relevant value principles derived from Chinese philosophical wisdom.
\end{itemize}

\begin{itemize}
\item We derive driving-relevant value principles from Confucian classics and organize them into retrievable, scenario-specific cases, and introduce PSSR for historical scene retrieval.

\item Closed-loop experiments across three traffic configurations show higher success rates, more collision-free steps, and restrained driving behavior under Confucian value guidance, with an efficiency trade-off.
\end{itemize}

\section{Related Work}

\subsection{LLMs for Autonomous Driving Decision-Making}

The reasoning and commonsense capabilities of LLMs have extended their use to sequential decision systems, as demonstrated by reasoning-and-acting, self-reflection, and skill-acquisition agents such as ReAct, Reflexion, and Voyager~\citep{yao2022react,shinn2023reflexion,wang2023voyager}. LLMs have also been applied to autonomous driving decisions. DriveLikeAHuman organizes decisions using a human-inspired cognitive structure~\citep{fu2024drive}. DiLu integrates reasoning, memory, and reflection into a knowledge-driven closed-loop driving framework~\citep{wen2024dilu}. GPT-Driver formulates planning as language-sequence generation~\citep{mao2023gpt}, while DriveGPT4 studies interpretable end-to-end autonomous driving with LLMs~\citep{xu2024drivegpt4}. LanguageMPC connects LLM reasoning to vehicle control through adaptive planning~\citep{sha2023languagempc}. LMDrive integrates multimodal sensor information with natural-language instructions for closed-loop driving~\citep{shao2024lmdrive}. LimSim++ provides a closed-loop evaluation platform~\citep{fu2024limsim++}. More recently, DriveGPT4-V2 further explores enhanced closed-loop autonomous driving with multimodal LLMs~\citep{xu2025drivegpt4}.

These methods primarily address language-based planning, commonsense reasoning, and the use of historical experience. DriveReg and KnowVal additionally incorporate traffic regulations or learned value preferences~\citep{cai2026driving,xia2026knowval}. However, how philosophical principles can be grounded in concrete driving scenarios to support decision-making remains underexplored.

\subsection{Ethical and Philosophical Values in Autonomous Driving}

Early autonomous-driving ethics mainly focused on collision dilemmas and harm allocation in unavoidable accidents. \citet{bonnefon2016social} identified a public-acceptance paradox for utilitarian autonomous vehicles, while \citet{awad2018moral} used the Moral Machine to reveal cross-cultural differences in ethical preferences. \citet{sui2024confucian} proposed a Confucian moral algorithm for trolley-type harm allocation, rather than LLM-based continuous closed-loop driving. However, dilemma studies provide an insufficient basis for policy and limited guidance for continuous real-world driving, while policy discussions address broader normative considerations~\citep{dewitt2019moral,eudg2020ethics}.

Technical approaches largely formalize traffic rules, safety principles, and ethical requirements as constraints, cost functions, or decision rules. The Responsibility-Sensitive Safety (RSS) model encodes safety responsibilities as proper-response rules~\citep{shalev2017formal}, while legal-rule models represent traffic regulations and exceptions as structured obligations~\citep{bhuiyan2023driving}. Although operationally clear, these methods provide limited support for broader value considerations.

LLMs offer an alternative means of incorporating regulations, social norms, and ethical knowledge expressed in natural language, with recent driving studies exploring regulatory compliance and preference alignment~\citep{cai2026driving,xia2026knowval}. Beyond driving, ConfuciusLM applies Confucian pedagogical principles through fine-tuning for primary education, whereas CyberConfucius retrieves classical Confucian texts for philosophical counseling~\citep{xu2026confuciuslm,zhao2025cyberconfucius}. FLAMES and CMoralEval benchmark Chinese values and moral norms for LLM evaluation~\citep{huang2024flames,yu2024cmoraleval}.

Existing studies have incorporated philosophical and culturally situated values into LLM training, retrieval, or evaluation, but their use in continuous driving decision-making remains underexplored. Rather than formalizing Confucian values as scalar rewards or optimization objectives, CPW-Drive represents them through natural-language principles and scenario-grounded historical cases to guide LLM-based closed-loop driving decisions at inference time.

\subsection{Retrieval-Augmented Decision-Making and Scene Similarity Matching}

In autonomous driving, retrieval-augmented methods use external knowledge or historical experience to support decision-making. Existing approaches commonly retrieve relevant knowledge or historical cases based on natural-language descriptions or embedding representations~\citep{cai2026driving,xia2026knowval,ye2025safedriverag,wen2024dilu,mao2023language}. However, semantically similar representations may correspond to driving scenes with substantially different spatial configurations and motion states.

To better characterize driving scenes themselves, recent studies have incorporated structured driving information, including BEV representations, scene graphs, trajectory graphs, and spatiotemporal structures, into scene retrieval~\citep{tang2025bev,tian2025query,cheng2024vetrass,zhao2026vladriver}. These methods typically measure scene similarity through learned scene representations or structural matching, allowing spatial relations, interactions, and trajectory information to be reflected in the retrieval process. In such approaches, scene similarity is generally determined through the resulting representations or structures rather than by an explicit similarity function defined directly over vehicle physical states.

Unlike the representation- or structure-based matching approaches above, PSSR defines scene similarity directly from vehicle physical states. It jointly considers the current spatial configuration and a short-term future spatial state estimated from current vehicle velocities, allowing the similarity calculation to reflect both spatial layout and motion trends.

\section{Methodology}

\begin{figure*}[t]
\centering
\includegraphics[width=0.98\textwidth,height=0.72\textheight,keepaspectratio]{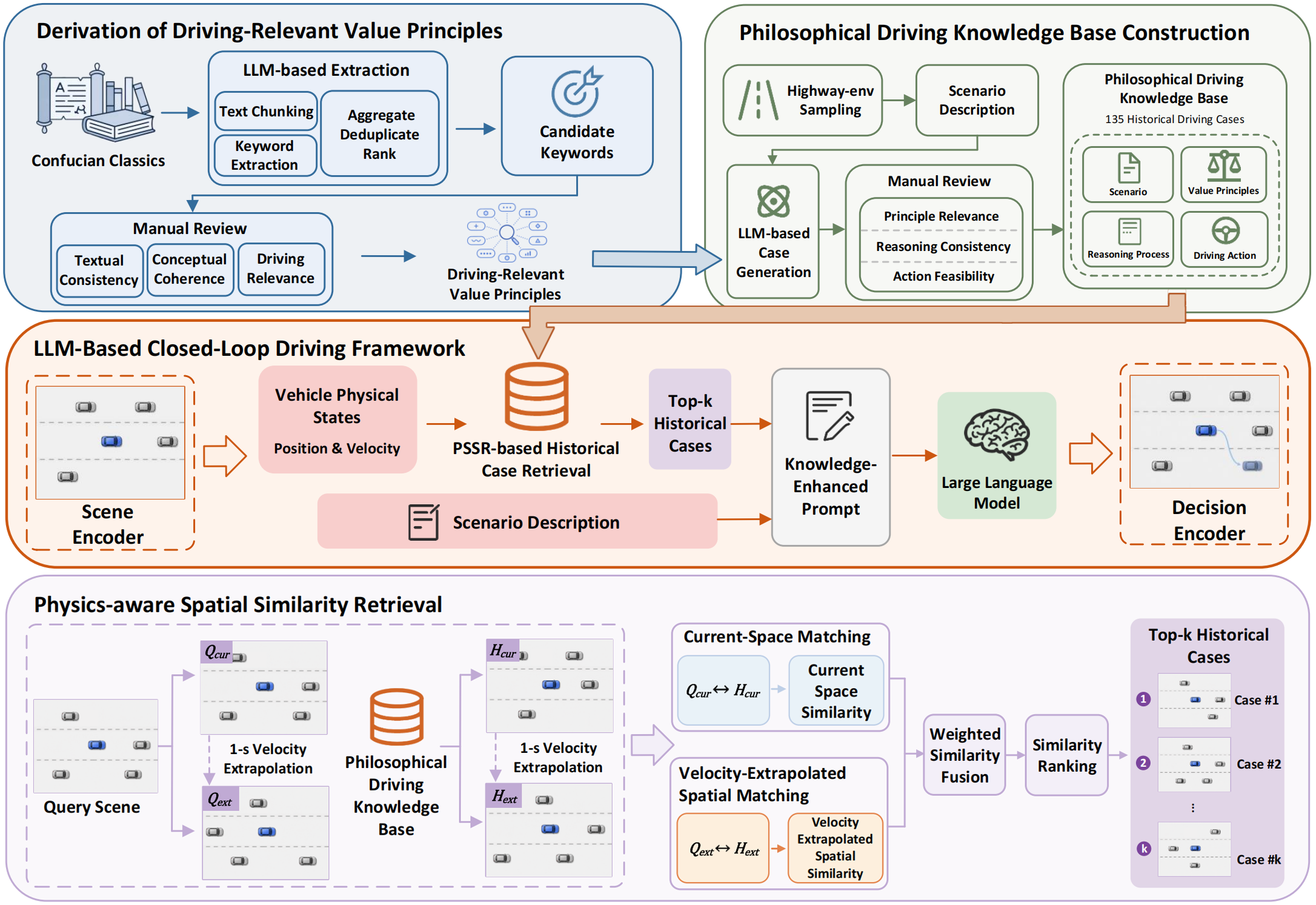}
\caption{System overview of CPW-Drive. Driving-relevant value principles are extracted and combined with scenario-specific driving cases to construct the Philosophical Driving Knowledge Base, from which PSSR retrieves relevant historical cases to support closed-loop LLM decision-making.}
\label{fig:framework}
\end{figure*}

Figure~\ref{fig:framework} illustrates the overall architecture of CPW-Drive, which consists of four main components: driving-relevant value principle construction, philosophical driving knowledge-base construction, historical scene retrieval, and closed-loop LLM decision-making.

\subsection{LLM-Based Closed-Loop Driving Framework}

CPW-Drive operates as a continuous observation, decision-making, and execution loop. At each step, the Scene Encoder converts the environment state into a structured natural-language Scenario Description for LLM reasoning. The system retains vehicle physical states for PSSR to retrieve relevant historical cases from the Philosophical Driving Knowledge Base. Each case contains four fields: scenario, value principles, reasoning process, and driving action.

The LLM reasons over the current Scenario Description, retrieved cases, and available driving actions to generate the current reasoning process and driving action. The Decision Encoder processes the generated decision for execution in the driving environment, and the updated environment state enters the next step.

The prompts used by CPW-Drive and an illustrative closed-loop decision example are provided in Appendix~\ref{app:prompt_example}.

\subsection{Derivation of Driving-Relevant Value Principles}

We use 131 Confucian classics as the knowledge source and divide the texts into chunks, from which an LLM extracts keywords that capture their core ideas. Keywords extracted across different classics and text chunks are then aggregated, deduplicated, and ranked by identification frequency to form a candidate keyword set. The resulting candidate set contained 40 keywords. The titles of the 131 Confucian classics are listed in Appendix~\ref{app:confucian_classics}.

Based on the candidate keyword set, we manually review, screen, and conceptually integrate the keywords according to their consistency with the source texts, conceptual overlap, and relevance to driving decision-making, consolidating similar or redundant terms and removing weakly relevant candidates. The resulting set is then subject to a final review by one of the authors, who holds a doctoral degree in Chinese philosophy. This process yields ten driving-relevant value principles: Wisdom, Righteousness, Respect, Adaptability, Benevolence, Courage, Moderation, Propriety, Empathy, and Trustworthiness.

These value principles provide high-level value considerations rather than directly prescribing driving actions. We further combine them with driving scenarios, reasoning processes, and actions to construct historical driving cases for the philosophical driving knowledge base.

\subsection{Philosophical Driving Knowledge Base Construction}

The Philosophical Driving Knowledge Base is organized as a set of historical driving cases. Each case contains four fields: scenario, value principles, reasoning process, and driving action, and can be represented as \(C=(S,V,R,A)\). Here, \(S\) describes the traffic state at the decision moment, including the positions, velocities, and accelerations of the ego vehicle and surrounding vehicles; \(V\) denotes the value principles relevant to the scenario; \(R\) records the corresponding reasoning process; and \(A\) specifies the resulting driving action. This structure preserves both the driving decision and its value-guided reasoning process.

To construct historical driving cases, we collect more than 3,000 candidate scenarios using different random seeds exclusively under the five-lane configuration with a vehicle density of 2.5 in Highway-env. The random seeds used for case construction do not overlap with those used in any subsequent evaluation experiments, and no other traffic configurations are used for case construction. We screen the scenarios based on vehicle spatial configurations, relative motion states, and driving constraints. Specifically, we consider the relative positions and lane relationships between the ego vehicle and surrounding vehicles, the relative velocities and time-to-collision of key vehicles, car-following distances, the ego vehicle’s lane position, and the gaps to vehicles ahead and behind in adjacent lanes. For scenarios with highly similar characteristics, we retain representative scenarios to reduce redundancy, while retaining scenarios with similar layouts but different conflict urgency or feasible driving options as distinct cases. This process yields 135 representative scenarios. For each scenario, we provide the scenario information and the full set of value principles derived in the preceding section to the LLM, which identifies relevant principles and generates the corresponding reasoning process and driving action to form an initial historical case.

The annotations of these 135 cases are then reviewed and revised by one of the authors, who holds a doctoral degree in Chinese philosophy. The review addresses three aspects: whether the selected value principles are consistent with their Confucian meanings and applicable to the scenario; whether the reasoning process is consistent with the scenario and selected principles while following conventional safe-driving reasoning; and whether the driving action is available in the current state, reasonable, feasible, and consistent with the reasoning. For annotations that do not meet these criteria, the corresponding value principles, reasoning process, or driving action are revised. The resulting knowledge base contains 135 historical driving cases for subsequent PSSR-based retrieval.

\subsection{Physics-aware Spatial Similarity Retrieval}
\label{sec:pssr}

PSSR is inspired by Smooth Overlap of Atomic Positions (SOAP), which was originally developed for comparing local atomic structures~\citep{bartok2013representing}. We adapt its continuous-density-overlap idea to ego-centered driving scenes, introduce anisotropic longitudinal and lateral spatial scales, and estimate short-term future vehicle positions from current velocities to capture both spatial structure and motion trends.

For the current scene \(A\) and a historical scene \(B\), both are represented in an ego-centered coordinate system as

\[
Z=\left(v_{\mathrm{ego},x}^{Z},
\left\{(x_i^{Z},y_i^{Z},v_{i,x}^{Z})\right\}_{i=1}^{N_Z}\right),
\qquad Z\in\{A,B\}.
\]

Here, \(Z\in\{A,B\}\) indexes the current or historical scene, \(x\) and \(y\) denote the longitudinal and lateral positions of surrounding vehicles relative to the ego vehicle, respectively, \(v_x\) denotes the longitudinal vehicle velocity, and \(v_{\mathrm{ego},x}\) denotes the ego-vehicle longitudinal velocity. The ego vehicle serves as the coordinate reference and is not included in the subsequent spatial density field, while its velocity is used to characterize the relative motion of surrounding vehicles. All historical scenes in the knowledge base contain surrounding vehicles. When no surrounding vehicle is present in the current scene, PSSR-based historical case retrieval is skipped; hence, the following similarity computation is defined only for non-empty vehicle sets.

In addition to the current spatial state, PSSR estimates a short-term future spatial state from the current vehicle velocities. For vehicle \(i\) in scene \(A\), the extrapolated position is defined as

\begin{equation}
\label{eq:velocity_extrapolation}
\hat{x}_i^{A}
=
x_i^{A}
+
\left(v_{i,x}^{A}-v_{\mathrm{ego},x}^{A}\right)\Delta t,
\qquad
\hat{y}_i^{A}=y_i^{A},
\end{equation}

where \(\Delta t=1.0\,\mathrm{s}\). The same extrapolation is applied to scene \(B\). Each scene therefore forms a current-space point set and a velocity-extrapolated point set:

\[
\left(P_{\mathrm{cur}}(A),P_{\mathrm{ext}}(A)\right)
=
\left(
\left\{(x_i^{A},y_i^{A})\right\}_{i=1}^{N_A},
\left\{(\hat{x}_i^{A},\hat{y}_i^{A})\right\}_{i=1}^{N_A}
\right),
\]

with analogous definitions for scene \(B\). By considering both the current vehicle layout and the short-term extrapolated state, PSSR captures spatial configuration together with motion trends in the subsequent similarity calculation. Compared with using the current spatial state alone, incorporating the extrapolated state accounts for relative velocity differences and can therefore retrieve cases with more closely matched motion conditions.

To characterize structural similarity between vehicles in road space, PSSR represents the position of each surrounding vehicle using a two-dimensional anisotropic Gaussian density:

\begin{equation}
\label{eq:anisotropic_density}
\rho_i(x,y)
=
\exp\left(
-\frac{(x-x_i)^2}{2\sigma_x^2}
-\frac{(y-y_i)^2}{2\sigma_y^2}
\right).
\end{equation}

Here, \(\sigma_x\) and \(\sigma_y\) control the longitudinal and lateral spatial scales, respectively. Anisotropic spatial scales are used because longitudinal following distances and lateral lane offsets have different physical meanings and scales. The overlap between the Gaussian densities of two vehicles is rescaled by the common factor \(\pi\sigma_x\sigma_y\) to define the vehicle-level spatial kernel:

\begin{equation}
\label{eq:vehicle_spatial_kernel}
\begin{aligned}
K_{\mathrm{space}}(i,j)
&=
\frac{1}{\pi\sigma_x\sigma_y}
\iint
\rho_i(x,y)\rho_j(x,y)\,dxdy\\
&=
\exp\left(
-\frac{(x_i-x_j)^2}{4\sigma_x^2}
-\frac{(y_i-y_j)^2}{4\sigma_y^2}
\right).
\end{aligned}
\end{equation}

We set \(\sigma_x=5\,\mathrm{m}\), and determine \(\sigma_y\approx1.1533\,\mathrm{m}\) by requiring the spatial kernel to decay to \(0.10\) at a lateral offset of one lane width, \(L=3.5\,\mathrm{m}\).

For two vehicle point sets \(X\) and \(Y\), the scene-level raw spatial overlap is defined as the sum of the spatial similarities over all vehicle pairs:

\begin{equation}
\label{eq:scene_spatial_overlap}
\begin{aligned}
&S_{\mathrm{space}}(X,Y)
=
\sum_{i=1}^{N_X}
\sum_{j=1}^{N_Y}
\\
&\quad \exp\left(
-\frac{(x_i^{X}-x_j^{Y})^2}{4\sigma_x^2}
-\frac{(y_i^{X}-y_j^{Y})^2}{4\sigma_y^2}
\right).
\end{aligned}
\end{equation}

Because the computation sums over all vehicle pairs across the two scenes, it does not require an explicit one-to-one vehicle correspondence. To enable comparisons between scenes with different numbers of vehicles, we use the normalized form

\begin{equation}
\label{eq:normalized_spatial_similarity}
K_{\mathrm{space}}^{0}(X,Y)
=
\frac{
S_{\mathrm{space}}(X,Y)
}{
\sqrt{
S_{\mathrm{space}}(X,X)
S_{\mathrm{space}}(Y,Y)
}
}.
\end{equation}

Here, \(S_{\mathrm{space}}(X,X)\) denotes the self-overlap of scene \(X\), which includes cross terms between distinct vehicles within the scene and therefore cannot be simplified to the number of vehicles.

The current-space similarity and velocity-extrapolated-space similarity can be written compactly as

\begin{equation}
\label{eq:state_specific_similarity}
K_s(A,B)
\!=\!
K_{\mathrm{space}}^{0}
\left(
P_s(A),
P_s(B)
\right),
\, s\in\{\mathrm{cur},\mathrm{ext}\}.
\end{equation}

The two similarities are combined through a weighted average:

\begin{equation}
\label{eq:final_pssr_similarity}
K(A,B)
=
w_{\mathrm{cur}}K_{\mathrm{cur}}(A,B)
+
w_{\mathrm{ext}}K_{\mathrm{ext}}(A,B).
\end{equation}

We set \(w_{\mathrm{cur}}=0.7\) and \(w_{\mathrm{ext}}=0.3\), giving greater weight to the current spatial layout while using the extrapolated state to complement short-term motion trends. Because the available samples for weight tuning are limited, we do not perform an extensive hyperparameter search and instead adopt this heuristic setting. The resulting similarity lies in \((0,1]\). Detailed derivations of the Gaussian-overlap kernel, the lateral-scale parameter, the normalized similarity, and its value range are provided in Appendix~\ref{app:kernel_derivation}.

After obtaining the final PSSR similarity \(K(A,B)\) between the current scene \(A\) and each historical scene \(B\) in the knowledge base, we rank the historical cases in descending order of similarity and select the Top-\(k\) most relevant cases as references for the current LLM decision, with \(k=3\) in the main experiments. Appendix~\ref{app:retrieval_case_analysis} further compares representative retrieval results from text-vector retrieval, current-position-only spatial retrieval, relative-distance retrieval, TTC-based retrieval, and full PSSR to analyze differences among similarity measures and the role of velocity-extrapolated states in scene matching.

\section{Experiments}

\subsection{Experimental Setup}

We evaluate CPW-Drive in the \texttt{highway-v0} environment of the Highway-env simulator~\citep{highway-env}. The discrete meta-action space contains five driving actions: change lane left, maintain the current state, change lane right, accelerate, and decelerate. Following the experimental settings of DiLu~\citep{wen2024dilu}, we use three traffic configurations with increasing traffic complexity: four lanes with density 2.0, five lanes with density 2.5, and five lanes with density 3.0.

For each configuration, we conduct 10 evaluation rounds, each comprising 20 runs with different random seeds. Each run contains 30 decision steps, and a trial is considered successful if the ego vehicle completes all 30 steps without a collision. Main and ablation results are reported as the mean and standard deviation across the 10 rounds. Knowledge-base and test scenarios are generated with non-overlapping random seeds. All main experiments use 135 human-screened historical cases, with the top three PSSR-ranked cases retrieved at each decision step as reasoning demonstrations.

We compare CPW-Drive with DriveLikeAHuman~\citep{fu2024drive}, DiLu~\citep{wen2024dilu}, and LimSim++~\citep{fu2024limsim++}. DriveLikeAHuman represents an LLM-based human-like driving decision method, while DiLu and LimSim++ represent LLM-based autonomous driving methods incorporating knowledge retrieval or simulation-based reasoning. All methods use DeepSeek-V4-Pro through the same API without additional fine-tuning. The baselines are reproduced from their publicly available implementations, retaining their original prompt templates and method-specific settings, except that the retrieval databases of DiLu and LimSim++ are replaced with a common case database constructed from the same 135 scenarios and containing only scenario descriptions and driving actions, to control for differences in case sources and database size.

We use four evaluation metrics: success rate (SR), collision-free steps (CFS), efficiency score (ES)~\citep{jiang2024koma}, and lane-change frequency (LCF). SR is the proportion of trials that complete all 30 decision steps without collision, and CFS is the number of steps completed before collision or termination. Following KoMA, ES measures driving efficiency based on the ratio of the ego vehicle's speed to the average speed of surrounding vehicles; it is set to 10 when the ego vehicle's speed reaches or exceeds that average and is otherwise scaled proportionally. LCF records the frequency of lateral maneuvers and is treated as a descriptive behavioral metric rather than a metric with a monotonic preference.

\subsection{Main Performance Analysis}

\begin{table*}[t]
\centering
\caption{Main closed-loop driving performance under different traffic settings. Best and second-best results for SR, CFS, and ES within each setting are shown in bold and underlined, respectively. LCF is reported as a descriptive behavioral metric and is not ranked.}
\label{tab:main_results}

\footnotesize
\setlength{\tabcolsep}{1.6pt}
\renewcommand{\arraystretch}{1.00}

% ===================== Panel (a) =====================
\textbf{(a) Success and safety metrics}

\begin{tabular}{lcccccc}
\toprule
& \multicolumn{2}{c}{4 lanes, density = 2.0}
& \multicolumn{2}{c}{5 lanes, density = 2.5}
& \multicolumn{2}{c}{5 lanes, density = 3.0} \\
\cmidrule(lr){2-3}
\cmidrule(lr){4-5}
\cmidrule(lr){6-7}

Model
& SR $\uparrow$ (\%) & CFS $\uparrow$
& SR $\uparrow$ (\%) & CFS $\uparrow$
& SR $\uparrow$ (\%) & CFS $\uparrow$ \\
\midrule

DriveLikeAHuman
& $51.5 \pm 5.8$
& $19.75 \pm 1.25$
& $50.5 \pm 10.1$
& $\underline{24.50 \pm 1.07}$
& $33.5 \pm 5.8$
& $12.90 \pm 1.74$ \\

DiLu
& $75.0 \pm 7.5$
& $22.90 \pm 2.66$
& $59.0 \pm 5.7$
& $22.15 \pm 1.34$
& $\underline{47.0 \pm 5.9}$
& $14.24 \pm 1.44$ \\

LimSim++
& $\underline{85.0 \pm 8.8}$
& $\underline{27.35 \pm 1.98}$
& $\underline{63.5 \pm 8.8}$
& $21.25 \pm 2.13$
& $37.5 \pm 7.9$
& $\underline{14.32 \pm 2.51}$ \\

\textbf{CPW-Drive (Ours)}
& $\mathbf{93.0 \pm 5.4}$
& $\mathbf{28.12 \pm 1.51}$
& $\mathbf{86.0 \pm 7.0}$
& $\mathbf{27.10 \pm 1.55}$
& $\mathbf{72.0 \pm 6.3}$
& $\mathbf{22.44 \pm 1.37}$ \\

\bottomrule
\end{tabular}

\vspace{3pt}

% ===================== Panel (b) =====================
\textbf{(b) Efficiency and Driving Behavior Metrics}

\begin{tabular}{lcccccc}
\toprule
& \multicolumn{2}{c}{4 lanes, density = 2.0}
& \multicolumn{2}{c}{5 lanes, density = 2.5}
& \multicolumn{2}{c}{5 lanes, density = 3.0} \\
\cmidrule(lr){2-3}
\cmidrule(lr){4-5}
\cmidrule(lr){6-7}

Model
& ES $\uparrow$ & LCF (\%)
& ES $\uparrow$ & LCF (\%)
& ES $\uparrow$ & LCF (\%) \\
\midrule

DriveLikeAHuman
& $\underline{9.38 \pm 0.16}$
& $14.73 \pm 1.74$
& $\mathbf{9.45 \pm 0.20}$
& $9.54 \pm 1.68$
& $\underline{9.20 \pm 0.18}$
& $11.86 \pm 2.13$ \\

DiLu
& $9.31 \pm 0.17$
& $4.29 \pm 0.94$
& $9.27 \pm 0.16$
& $5.73 \pm 0.73$
& $9.17 \pm 0.20$
& $6.58 \pm 1.06$ \\

LimSim++
& $\mathbf{9.49 \pm 0.11}$
& $8.38 \pm 1.43$
& $\underline{9.42 \pm 0.14}$
& $8.21 \pm 1.27$
& $\mathbf{9.21 \pm 0.14}$
& $8.41 \pm 0.98$ \\

\textbf{CPW-Drive (Ours)}
& $8.38 \pm 0.13$
& $4.93 \pm 1.68$
& $8.32 \pm 0.11$
& $5.49 \pm 0.69$
& $8.47 \pm 0.25$
& $6.75 \pm 1.24$ \\

\bottomrule
\end{tabular}

\end{table*}

Table~\ref{tab:main_results} shows that CPW-Drive achieves the highest SR and CFS across all three traffic configurations. Compared with the strongest SR baseline in each configuration, its SR improves by 8.0, 22.5, and 25.0 percentage points, respectively. Its LCF also remains generally low, while its ES is lower than the best baseline in all three configurations. Overall, these results exhibit a performance pattern that places greater emphasis on safety and restraint, which is also consistent with the more restrained and accommodating driving tendency reflected in Confucian thought. A complementary third-party evaluation of the Confucian value conformity of CPW-Drive's reasoning and decisions is provided in Appendix~\ref{app:confucian_evaluation}. The following ablation study further examines the roles of retrieval, philosophical knowledge, and PSSR, and tests whether this performance pattern can be explained by generic conservative driving prompts.

\subsection{Ablation Study}

We evaluate three ablations and two controls to examine the roles of historical-case retrieval, PSSR, philosophical content, and general safety or conservative-driving guidance. \textbf{No RAG} removes historical-case retrieval; \textbf{Text RAG} replaces PSSR with text-vector retrieval using the same database and retrieval count; and \textbf{RAG w/o Philosophy} retains PSSR and the same historical scenarios but excludes value principles, generating reasoning and actions solely from scene information. \textbf{Safety Rules RAG} replaces Confucian principles with general traffic safety rules while keeping the framework, retrieval method, case count, and review criteria as consistent as possible, to compare the effects of the two types of guidance on driving decisions. \textbf{Conservative Prompt} combines non-philosophical historical cases with generic conservative-driving instructions. The relevant prompts are provided in Appendix~\ref{app:prompt_example}.

\begin{table*}[t]
\centering
\caption{Ablation results under different traffic settings. Best and second-best results for SR, CFS, and ES within each setting are shown in bold and underlined, respectively. LCF is reported as a descriptive behavioral metric and is not ranked.}
\label{tab:ablation}

\footnotesize
\setlength{\tabcolsep}{1.6pt}
\renewcommand{\arraystretch}{1.00}

% ===================== Panel (a) =====================
\textbf{(a) Success and safety metrics}

\begin{tabular}{lcccccc}
\toprule
& \multicolumn{2}{c}{4 lanes, density = 2.0}
& \multicolumn{2}{c}{5 lanes, density = 2.5}
& \multicolumn{2}{c}{5 lanes, density = 3.0} \\
\cmidrule(lr){2-3}
\cmidrule(lr){4-5}
\cmidrule(lr){6-7}
Method
& SR $\uparrow$ (\%) & CFS $\uparrow$
& SR $\uparrow$ (\%) & CFS $\uparrow$
& SR $\uparrow$ (\%) & CFS $\uparrow$ \\
\midrule

\textbf{CPW-Drive}
& $\mathbf{93.0 \pm 5.4}$
& $\underline{28.12 \pm 1.51}$
& $\mathbf{86.0 \pm 7.0}$
& $\mathbf{27.10 \pm 1.55}$
& $\mathbf{72.0 \pm 6.3}$
& $\mathbf{22.44 \pm 1.37}$ \\

No RAG
& $84.0 \pm 4.6$
& $25.85 \pm 1.22$
& $70.0 \pm 5.8$
& $24.10 \pm 1.27$
& $57.0 \pm 6.7$
& $18.05 \pm 1.69$ \\

Text RAG
& $84.5 \pm 6.0$
& $25.55 \pm 1.82$
& $74.5 \pm 5.5$
& $24.70 \pm 1.45$
& $61.0 \pm 7.7$
& $18.60 \pm 1.91$ \\

RAG w/o Philosophy
& $85.5 \pm 5.0$
& $26.45 \pm 1.08$
& $\underline{77.0 \pm 6.3}$
& $\underline{25.25 \pm 1.32}$
& $\underline{67.0 \pm 5.9}$
& $\underline{20.35 \pm 1.78}$ \\

Safety Rules RAG
& $\underline{88.0 \pm 6.8}$
& $27.25 \pm 1.12$
& $72.0 \pm 10.0$
& $22.12 \pm 1.51$
& $44.0 \pm 11.1$
& $14.56 \pm 2.23$ \\

Conservative Prompt
& $\mathbf{93.0 \pm 2.6}$
& $\mathbf{28.33 \pm 0.64}$
& $76.0 \pm 9.1$
& $23.86 \pm 3.09$
& $55.0 \pm 11.1$
& $18.88 \pm 2.98$ \\

\bottomrule
\end{tabular}

\vspace{3pt}

% ===================== Panel (b) =====================
\textbf{(b) Efficiency and Driving Behavior Metrics}

\begin{tabular}{lcccccc}
\toprule
& \multicolumn{2}{c}{4 lanes, density = 2.0}
& \multicolumn{2}{c}{5 lanes, density = 2.5}
& \multicolumn{2}{c}{5 lanes, density = 3.0} \\
\cmidrule(lr){2-3}
\cmidrule(lr){4-5}
\cmidrule(lr){6-7}
Method
& ES $\uparrow$ & LCF (\%)
& ES $\uparrow$ & LCF (\%)
& ES $\uparrow$ & LCF (\%) \\
\midrule

\textbf{CPW-Drive}
& $8.38 \pm 0.13$
& $4.93 \pm 1.68$
& $8.32 \pm 0.11$
& $5.49 \pm 0.69$
& $8.47 \pm 0.25$
& $6.75 \pm 1.24$ \\

No RAG
& $8.43 \pm 0.21$
& $7.55 \pm 1.76$
& $8.57 \pm 0.31$
& $7.70 \pm 1.49$
& $8.48 \pm 0.25$
& $9.35 \pm 1.84$ \\

Text RAG
& $\underline{8.67 \pm 0.09}$
& $7.15 \pm 1.61$
& $\underline{8.63 \pm 0.19}$
& $8.35 \pm 1.42$
& $\underline{8.83 \pm 0.19}$
& $8.95 \pm 1.65$ \\

RAG w/o Philosophy
& $\mathbf{8.85 \pm 0.23}$
& $8.60 \pm 1.87$
& $\mathbf{9.02 \pm 0.10}$
& $9.75 \pm 1.66$
& $\mathbf{9.08 \pm 0.13}$
& $10.35 \pm 1.92$ \\

Safety Rules RAG
& $8.35 \pm 0.36$
& $4.13 \pm 1.30$
& $8.47 \pm 0.51$
& $5.07 \pm 1.38$
& $8.82 \pm 0.24$
& $6.88 \pm 1.37$ \\

Conservative Prompt
& $5.72 \pm 0.26$
& $1.28 \pm 0.16$
& $5.75 \pm 0.36$
& $1.80 \pm 0.17$
& $6.20 \pm 0.38$
& $2.73 \pm 0.50$ \\

\bottomrule
\end{tabular}

\end{table*}

Retrieval ablations support complementary roles for historical cases and PSSR. Relative to CPW-Drive, \textbf{No RAG} reduces SR and CFS and increases LCF, showing that retrieved cases support current-scene decisions. \textbf{Text RAG} also reduces SR and CFS, indicating that PSSR retrieves better-matched cases by modeling spatial relations and motion states.

\begin{figure*}[t]
\centering
\includegraphics[width=0.98\textwidth,height=0.72\textheight,keepaspectratio]{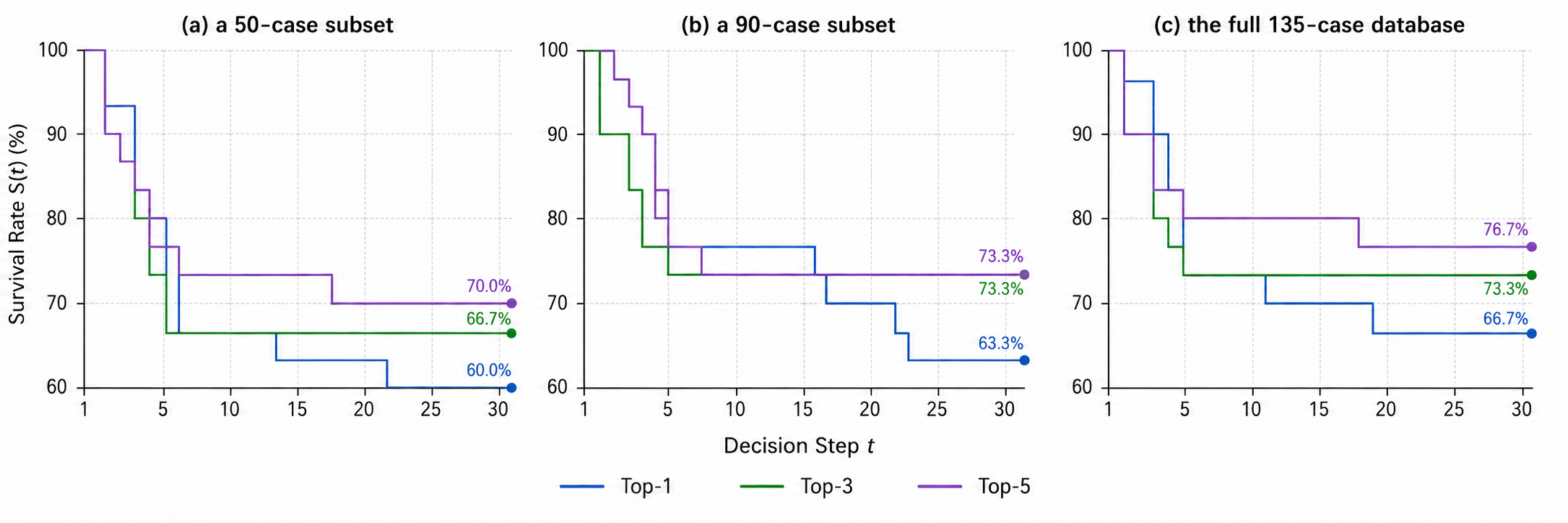}
\caption{Safe-step survival curves under different database sizes and Top-$k$ settings. Panels (a)--(c) correspond to 50, 90, and 135 historical cases, respectively.}
\label{fig:sensitivity}
\end{figure*}

The philosophy-related comparisons reveal a safety-efficiency trade-off. Relative to CPW-Drive, \textbf{RAG w/o Philosophy} yields lower SR and CFS but higher ES and LCF across all configurations, indicating that philosophical content contributes to safer, more restrained behavior at an efficiency cost. \textbf{Safety Rules RAG} likewise achieves lower SR and CFS than CPW-Drive across all configurations, with the SR gap widening from 5 to 28 percentage points as traffic complexity increases. Although it achieves slightly higher ES in the two higher-density configurations, it does not attain comparable safety performance. These results indicate that CPW-Drive's safety improvements cannot be attributed solely to general traffic safety guidance, supporting an additional role for Confucian value principles in scenario-grounded driving reasoning. \textbf{Conservative Prompt} produces the lowest LCF but incurs greater efficiency loss and degrades more sharply as traffic complexity increases. Thus, CPW-Drive's performance cannot be explained by generic conservatism alone, and excessive restrictions on proactive maneuvers, such as lane changes and acceleration, may reduce adaptability to complex traffic.

To examine retrieval-database size and top-$k$ settings, we conduct a sensitivity analysis in the highest-density environment. Figure~\ref{fig:sensitivity} reports 30 random seeds, with at most 30 decision steps per run. We compare databases containing 50, 90, or all 135 historical cases under top-1, top-3, and top-5 retrieval.

The sensitivity results show that overall survival performance improves as the number of historical cases increases, indicating that broader scenario coverage helps retrieve more relevant reference cases. Performance also generally improves as Top-\(k\) increases; however, while Top-3 and Top-5 achieve similar final performance, Top-5 exhibits a higher survival rate at earlier stages.

\subsection{Cross-Scenario Transfer Evaluation}

To evaluate the cross-scenario transfer capability of CPW-Drive, we extend its evaluation from straight-highway driving to the highway on-ramp merging scenario used in KoMA and directly reuse the 135 historical driving cases constructed under the original straight-highway configuration to assess the applicability of the existing knowledge base to different road layouts and interaction tasks. The scenario includes two vehicles controlled by CPW-Drive, with the mainline vehicle navigating safely while interacting with surrounding vehicles and the ramp vehicle merging safely onto the mainline. Each vehicle retrieves reference cases from the same historical case database based on its own current scene and independently generates driving decisions, without communication between the two vehicles.

We compare CPW-Drive with KoMA using the same four metrics as in the main experiments. Evaluation likewise comprises 10 rounds of 20 independent runs with different random seeds, with results reported as the mean and standard deviation across the 10 rounds. Each run lasts at most 15 decision steps and is considered successful if the ramp vehicle merges onto the mainline within this limit and neither controlled vehicle experiences a collision. CFS is measured as the number of decision steps completed when the overall scenario terminates, while ES and LCF are each averaged over the two controlled vehicles.

\begin{table}[t]
\centering
\caption{Cross-scenario transfer performance in highway on-ramp merging. Best results for SR, CFS, and ES are shown in bold.}
\label{tab:transfer_results}
\scriptsize
\setlength{\tabcolsep}{2pt}
\begin{tabular}{lcccc}
\toprule
Method & SR $\uparrow$ (\%) & CFS $\uparrow$ & ES $\uparrow$ & LCF (\%) \\
\midrule
KoMA & $68.0 \pm 7.6$ & $11.24 \pm 1.47$ & $\mathbf{9.47 \pm 0.09}$ & $17.04 \pm 4.73$ \\
\textbf{CPW-Drive} & $\mathbf{82.0 \pm 10.4}$ & $\mathbf{13.71 \pm 1.12}$ & $9.41 \pm 0.11$ & $15.78 \pm 1.10$ \\
\bottomrule
\end{tabular}
\end{table}

Table~\ref{tab:transfer_results} shows that CPW-Drive can transfer to on-ramp merging using only the original straight-highway case database. Compared with KoMA, it improves SR by 14 percentage points and achieves more collision-free steps, with only a slight decrease in driving efficiency and a lower lane-change frequency. These results suggest that the altruistic orientation of Confucian thought helps vehicles account for other vehicles' passage needs when making independent decisions, thereby promoting coordination among vehicles and improving overall merging performance.

\section{Conclusion}

This paper presents CPW-Drive, a closed-loop RAG framework that integrates Chinese philosophical wisdom into LLM-based autonomous driving decision-making. The framework derives driving-relevant value principles from Confucian classics and organizes them together with specific driving scenarios, reasoning processes, and actions in the Philosophical Driving Knowledge Base, while PSSR retrieves relevant historical cases according to vehicle spatial structure and motion trends. Across three simulated traffic configurations, CPW-Drive achieves the highest SR and CFS and maintains a relatively low LCF, while exhibiting lower ES.

The current study is limited by a knowledge base containing 135 historical cases and relatively short 30-step trials. Future work will further evaluate the framework in more diverse driving environments and with larger knowledge bases.

\subsection*{LLM Usage Statement}

OpenAI's ChatGPT was used to assist with language polishing and improving the clarity of author-written manuscript content, as well as refining author-developed code. Additionally, the DeepSeek API was employed to generate model responses as part of the experimental evaluation. All AI-assisted modifications were reviewed and verified by the authors, who retain full responsibility for the research methodology, experimental results, and scientific conclusions.

{\small
\bibliographystyle{ieeenat_fullname}
\bibliography{references}

@article{kiran2021deep,
  title={{Deep Reinforcement Learning for Autonomous Driving: A Survey}},
  author={Kiran, B Ravi and Sobh, Ibrahim and Talpaert, Victor and Mannion, Patrick and Al Sallab, Ahmad A and Yogamani, Senthil and P{\'e}rez, Patrick},
  journal={IEEE Transactions on Intelligent Transportation Systems},
  volume={23},
  number={6},
  pages={4909--4926},
  year={2022},
  publisher={IEEE},
  doi={10.1109/TITS.2021.3054625}
}

@article{zhu2021survey,
  title={{A Survey of Deep RL and IL for Autonomous Driving Policy Learning}},
  author={Zhu, Zeyu and Zhao, Huijing},
  journal={IEEE Transactions on Intelligent Transportation Systems},
  volume={23},
  number={9},
  pages={14043--14065},
  year={2022},
  publisher={IEEE},
  doi={10.1109/TITS.2021.3134702}
}

@inproceedings{tian2024tokenize,
  title={{Tokenize the World into Object-Level Knowledge to Address Long-Tail Events in Autonomous Driving}},
  author={Tian, Thomas and Li, Boyi and Weng, Xinshuo and Chen, Yuxiao and Schmerling, Edward and Wang, Yue and Ivanovic, Boris and Pavone, Marco},
  booktitle={Proceedings of the 8th Conference on Robot Learning},
  series={Proceedings of Machine Learning Research},
  volume={270},
  pages={3656--3673},
  year={2025},
  publisher={PMLR},
  url={https://proceedings.mlr.press/v270/tian25b.html}
}

@inproceedings{mao2023language,
  title={{A Language Agent for Autonomous Driving}},
  author={Mao, Jiageng and Ye, Junjie and Qian, Yuxi and Pavone, Marco and Wang, Yue},
  booktitle={First Conference on Language Modeling},
  year={2024},
  url={https://openreview.net/forum?id=UPE6WYE8vg}
}

@inproceedings{wen2024dilu,
  title={{DiLu: A Knowledge-Driven Approach to Autonomous Driving with Large Language Models}},
  author={Wen, Licheng and Fu, Daocheng and Li, Xin and Cai, Xinyu and Ma, Tao and Cai, Pinlong and Dou, Min and Shi, Botian and He, Liang and Qiao, Yu},
  booktitle={12th International Conference on Learning Representations, ICLR 2024},
  year={2024},
  url={https://openreview.net/forum?id=OqTMUPuLuC}
}

@article{mao2023gpt,
  title={{GPT-Driver: Learning to Drive with GPT}},
  author={Mao, Jiageng and Qian, Yuxi and Ye, Junjie and Zhao, Hang and Wang, Yue},
  journal={arXiv preprint arXiv:2310.01415},
  year={2023},
  doi={10.48550/arXiv.2310.01415},
  url={https://arxiv.org/abs/2310.01415}
}

@article{sha2023languagempc,
  title={{LanguageMPC: Large Language Models as Decision Makers for Autonomous Driving}},
  author={Sha, Hao and Mu, Yao and Jiang, Yuxuan and Chen, Li and Xu, Chenfeng and Luo, Ping and Li, Shengbo Eben and Tomizuka, Masayoshi and Zhan, Wei and Ding, Mingyu},
  journal={arXiv preprint arXiv:2310.03026},
  year={2023},
  doi={10.48550/arXiv.2310.03026},
  url={https://arxiv.org/abs/2310.03026}
}

@inproceedings{yao2022react,
  title={{ReAct: Synergizing Reasoning and Acting in Language Models}},
  author={Yao, Shunyu and Zhao, Jeffrey and Yu, Dian and Du, Nan and Shafran, Izhak and Narasimhan, Karthik R and Cao, Yuan},
  booktitle={The Eleventh International Conference on Learning Representations},
  year={2023},
  url={https://openreview.net/forum?id=WE_vluYUL-X}
}

@article{wang2023voyager,
  title={{Voyager: An Open-Ended Embodied Agent with Large Language Models}},
  author={Wang, Guanzhi and Xie, Yuqi and Jiang, Yunfan and Mandlekar, Ajay and Xiao, Chaowei and Zhu, Yuke and Fan, Linxi and Anandkumar, Anima},
  journal={Transactions on Machine Learning Research},
  year={2024},
  url={https://openreview.net/forum?id=ehfRiF0R3a}
}

@inproceedings{shinn2023reflexion,
  title={{Reflexion: Language Agents with Verbal Reinforcement Learning}},
  author={Shinn, Noah and Cassano, Federico and Gopinath, Ashwin and Narasimhan, Karthik and Yao, Shunyu},
  booktitle={Advances in Neural Information Processing Systems},
  volume={36},
  pages={8634--8652},
  year={2023},
  doi={10.52202/075280-0377}
}

@inproceedings{shao2024lmdrive,
  title={{LMDrive: Closed-Loop End-to-End Driving with Large Language Models}},
  author={Shao, Hao and Hu, Yuxuan and Wang, Letian and Song, Guanglu and Waslander, Steven L and Liu, Yu and Li, Hongsheng},
  booktitle={Proceedings of the IEEE/CVF Conference on Computer Vision and Pattern Recognition (CVPR)},
  pages={15120--15130},
  year={2024},
  doi={10.1109/CVPR52733.2024.01432}
}

@inproceedings{fu2024drive,
  title={{Drive Like a Human: Rethinking Autonomous Driving with Large Language Models}},
  author={Fu, Daocheng and Li, Xin and Wen, Licheng and Dou, Min and Cai, Pinlong and Shi, Botian and Qiao, Yu},
  booktitle={2024 IEEE/CVF Winter Conference on Applications of Computer Vision Workshops (WACVW)},
  pages={910--919},
  year={2024},
  organization={IEEE},
  doi={10.1109/WACVW60836.2024.00102}
}

@inproceedings{fu2024limsim++,
  title={{LimSim++: A Closed-Loop Platform for Deploying Multimodal LLMs in Autonomous Driving}},
  author={Fu, Daocheng and Lei, Wenjie and Wen, Licheng and Cai, Pinlong and Mao, Song and Dou, Min and Shi, Botian and Qiao, Yu},
  booktitle={2024 IEEE Intelligent Vehicles Symposium (IV)},
  pages={1084--1090},
  year={2024},
  organization={IEEE},
  doi={10.1109/IV55156.2024.10588848}
}

@article{xu2024drivegpt4,
  title={{DriveGPT4: Interpretable End-to-End Autonomous Driving via Large Language Model}},
  author={Xu, Zhenhua and Zhang, Yujia and Xie, Enze and Zhao, Zhen and Guo, Yong and Wong, Kwan-Yee K and Li, Zhenguo and Zhao, Hengshuang},
  journal={IEEE Robotics and Automation Letters},
  volume={9},
  number={10},
  pages={8186--8193},
  year={2024},
  publisher={IEEE},
  doi={10.1109/LRA.2024.3440097}
}

@inproceedings{xu2025drivegpt4,
  title={{DriveGPT4-V2: Harnessing Large Language Model Capabilities for Enhanced Closed-Loop Autonomous Driving}},
  author={Xu, Zhenhua and Bai, Yan and Zhang, Yujia and Li, Zhuoling and Xia, Fei and Wong, Kwan-Yee K and Wang, Jianqiang and Zhao, Hengshuang},
  booktitle={Proceedings of the IEEE/CVF Conference on Computer Vision and Pattern Recognition (CVPR)},
  pages={17261--17270},
  year={2025},
  doi={10.1109/CVPR52734.2025.01609}
}

@misc{highway-env,
  author = {Leurent, Edouard},
  title = {{An Environment for Autonomous Driving Decision-Making}},
  year = {2018},
  publisher = {GitHub},
  journal = {GitHub repository},
  howpublished = {\url{https://github.com/eleurent/highway-env}},
}

@article{jiang2024koma,
  title={{KoMA: Knowledge-Driven Multi-Agent Framework for Autonomous Driving with Large Language Models}},
  author={Jiang, Kemou and Cai, Xuan and Cui, Zhiyong and Li, Aoyong and Ren, Yilong and Yu, Haiyang and Yang, Hao and Fu, Daocheng and Wen, Licheng and Cai, Pinlong},
  journal={IEEE Transactions on Intelligent Vehicles},
  volume={10},
  number={10},
  pages={4655--4668},
  year={2025},
  publisher={IEEE},
  doi={10.1109/TIV.2024.3488793}
}

@article{bartok2013representing,
  author  = {Bart{\'o}k, A. P. and Kondor, R. and Cs{\'a}nyi, G.},
  title   = {{On Representing Chemical Environments}},
  journal = {Physical Review B},
  volume  = {87},
  number  = {18},
  pages   = {184115},
  year    = {2013},
  doi     = {10.1103/PhysRevB.87.184115}
}

@article{bonnefon2016social,
  author  = {Bonnefon, Jean-Fran{\c{c}}ois and Shariff, Azim and Rahwan, Iyad},
  title   = {{The Social Dilemma of Autonomous Vehicles}},
  journal = {Science},
  volume  = {352},
  number  = {6293},
  pages   = {1573--1576},
  year    = {2016},
  doi     = {10.1126/science.aaf2654}
}

@article{awad2018moral,
  author  = {Awad, Edmond and Dsouza, Sohan and Kim, Richard and Schulz, Jonathan and Henrich, Joseph and Shariff, Azim and Bonnefon, Jean-Fran{\c{c}}ois and Rahwan, Iyad},
  title   = {{The Moral Machine Experiment}},
  journal = {Nature},
  volume  = {563},
  number  = {7729},
  pages   = {59--64},
  year    = {2018},
  doi     = {10.1038/s41586-018-0637-6}
}

@article{sui2024confucian,
  author  = {Sui, Tingting and Sunday Gr{\`e}ve, Sebastian},
  title   = {{A Confucian Algorithm for Autonomous Vehicles}},
  journal = {Science and Engineering Ethics},
  volume  = {30},
  number  = {6},
  pages   = {52},
  year    = {2024},
  doi     = {10.1007/s11948-024-00514-z}
}

@article{dewitt2019moral,
  title   = {{`Moral Machine' Experiment Is No Basis for Policymaking}},
  author  = {Dewitt, Barry and Fischhoff, Baruch and Sahlin, Nils-Eric},
  journal = {Nature},
  volume  = {567},
  number  = {7746},
  pages   = {31},
  year    = {2019},
  doi     = {10.1038/d41586-019-00766-x}
}

@techreport{eudg2020ethics,
  title       = {{Ethics of Connected and Automated Vehicles: Recommendations on Road Safety, Privacy, Fairness, Explainability and Responsibility}},
  author      = {{Horizon 2020 Commission Expert Group to advise on specific ethical issues raised by driverless mobility (E03659)}},
  institution = {Publications Office of the European Union},
  address     = {Luxembourg},
  year        = {2020},
  isbn        = {978-92-76-17867-5},
  doi         = {10.2777/035239},
  url         = {https://op.europa.eu/en/publication-detail/-/publication/89624e2c-f98c-11ea-b44f-01aa75ed71a1/language-en/}
}

@article{shalev2017formal,
  title   = {{On a Formal Model of Safe and Scalable Self-Driving Cars}},
  author  = {Shalev-Shwartz, Shai and Shammah, Shaked and Shashua, Amnon},
  journal = {arXiv preprint arXiv:1708.06374},
  year    = {2017},
  doi     = {10.48550/arXiv.1708.06374},
  url     = {https://arxiv.org/abs/1708.06374}
}

@article{bhuiyan2023driving,
  title     = {{Driving Decision Making of Autonomous Vehicle According to Queensland Overtaking Traffic Rules}},
  author    = {Bhuiyan, Hanif and Governatori, Guido and Rakotonirainy, Andry and Wong, Meng Weng and Mahajan, Avishkar},
  journal   = {The Review of Socionetwork Strategies},
  volume    = {17},
  number    = {2},
  pages     = {233--254},
  year      = {2023},
  publisher = {Springer},
  doi       = {10.1007/s12626-023-00147-x}
}

@article{cai2026driving,
  title     = {{Driving with Regulation: Trustworthy and Interpretable Decision-Making for Autonomous Driving with Retrieval-Augmented Reasoning}},
  author    = {Cai, Tianhui and Liu, Yifan and Zhou, Zewei and Ma, Haoxuan and Zhao, Seth Z and Wu, Zhiwen and Han, Xu and Huang, Zhiyu and Ma, Jiaqi},
  journal   = {Proceedings of the AAAI Conference on Artificial Intelligence},
  volume    = {40},
  number    = {45},
  pages     = {38287--38295},
  year      = {2026},
  doi       = {10.1609/aaai.v40i45.41168}
}

@inproceedings{xia2026knowval,
  title     = {{KnowVal: A Knowledge-Augmented and Value-Guided Autonomous Driving System}},
  author    = {Xia, Zhongyu and Chen, Wenhao and Wang, Yongtao and Yang, Ming-Hsuan},
  booktitle = {Proceedings of the IEEE/CVF Conference on Computer Vision and Pattern Recognition (CVPR)},
  pages     = {3740--3749},
  year      = {2026}
}

@inproceedings{ye2025safedriverag,
  title     = {{SafeDriveRAG: Towards Safe Autonomous Driving with Knowledge Graph-Based Retrieval-Augmented Generation}},
  author    = {Ye, Hao and Qi, Mengshi and Liu, Zhaohong and Liu, Liang and Ma, Huadong},
  booktitle = {Proceedings of the 33rd ACM International Conference on Multimedia},
  pages     = {11170--11178},
  year      = {2025},
  doi       = {10.1145/3746027.3755868}
}

@article{tang2025bev,
  title     = {{BEV-TSR: Text-Scene Retrieval in BEV Space for Autonomous Driving}},
  author    = {Tang, Tao and Wei, Dafeng and Jia, Zhengyu and Gao, Tian and Cai, Changwei and Hou, Chengkai and Jia, Peng and Zhan, Kun and Sun, Haiyang and Fan, Jingchen and Zhao, Yixing and Liang, Xiaodan and Lang, Xianpeng and Wang, Yang},
  journal   = {Proceedings of the AAAI Conference on Artificial Intelligence},
  volume    = {39},
  number    = {7},
  pages     = {7275--7283},
  year      = {2025},
  doi       = {10.1609/aaai.v39i7.32782}
}

@article{tian2025query,
  title     = {{Query by Example: Semantic Traffic Scene Retrieval Using LLM-Based Scene Graph Representation}},
  author    = {Tian, Yafu and Carballo, Alexander and Li, Ruifeng and Thompson, Simon and Takeda, Kazuya},
  journal   = {Sensors},
  volume    = {25},
  number    = {8},
  pages     = {2546},
  year      = {2025},
  publisher = {MDPI},
  doi       = {10.3390/s25082546}
}

@article{cheng2024vetrass,
  title   = {{VeTraSS: Vehicle Trajectory Similarity Search Through Graph Modeling and Representation Learning}},
  author  = {Cheng, Ming and Zhang, Bowen and Wang, Ziyu and Zhou, Ziyi and Feng, Weiqi and Lyu, Yi and Diao, Xingjian},
  journal = {arXiv preprint arXiv:2404.08021},
  year    = {2024},
  doi     = {10.48550/arXiv.2404.08021},
  url     = {https://arxiv.org/abs/2404.08021}
}

@article{zhao2026vladriver,
  title   = {{VLADriver-RAG: Retrieval-Augmented Vision-Language-Action Models for Autonomous Driving}},
  author  = {Zhao, Rui and Hu, Haofeng and Gao, Zhenhai and Liu, Jiaqiao and Gao, Fei},
  journal = {arXiv preprint arXiv:2605.08133},
  year    = {2026},
  doi     = {10.48550/arXiv.2605.08133},
  url     = {https://arxiv.org/abs/2605.08133}
}

@article{xu2026confuciuslm,
  title   = {{ConfuciusLM: Exploring Confucian Personalized Instruction for Primary Learners With Large Language Models}},
  author  = {Xu, Xin and Lu, Xinya and Yan, Yusi and Tian, Mao and Wang, Jianan},
  journal = {IEEE Transactions on Learning Technologies},
  volume  = {19},
  pages   = {433--446},
  year    = {2026},
  doi     = {10.1109/TLT.2026.3688630}
}

@inproceedings{zhao2025cyberconfucius,
  title     = {{CyberConfucius: An Interactive Confucian Philosophical Counseling System for Mental Health}},
  author    = {Zhao, Liang and Chen, Bokai and Zheng, Weiwei and Zhou, Lingyan and Ding, Xiaojun},
  booktitle = {Companion of the 2025 ACM International Joint Conference on Pervasive and Ubiquitous Computing},
  pages     = {343--347},
  year      = {2025},
  doi       = {10.1145/3714394.3754364}
}

@inproceedings{huang2024flames,
  title     = {{FLAMES: Benchmarking Value Alignment of LLMs in Chinese}},
  author    = {Huang, Kexin and Liu, Xiangyang and Guo, Qianyu and Sun, Tianxiang and Sun, Jiawei and Wang, Yaru and Zhou, Zeyang and Wang, Yixu and Teng, Yan and Qiu, Xipeng and Wang, Yingchun and Lin, Dahua},
  booktitle = {Proceedings of the 2024 Conference of the North American Chapter of the Association for Computational Linguistics: Human Language Technologies},
  pages     = {4551--4591},
  year      = {2024},
  doi       = {10.18653/v1/2024.naacl-long.256}
}

@inproceedings{yu2024cmoraleval,
  title     = {{CMoralEval: A Moral Evaluation Benchmark for Chinese Large Language Models}},
  author    = {Yu, Linhao and Leng, Yongqi and Huang, Yufei and Wu, Shang and Liu, Haixin and Ji, Xinmeng and Zhao, Jiahui and Song, Jinwang and Cui, Tingting and Cheng, Xiaoqing and Liu, Tao and Xiong, Deyi},
  booktitle = {Findings of the Association for Computational Linguistics: ACL 2024},
  pages     = {11817--11837},
  year      = {2024},
  doi       = {10.18653/v1/2024.findings-acl.703}
}
}

\clearpage

\appendix
\section{Confucian Classical Works}
\label{app:confucian_classics}

This appendix lists the 131 Confucian and related classical works used to derive the driving-relevant value principles. Each entry provides the title in Hanyu Pinyin followed by its original Chinese title. Where a well-established English title exists, it is also included; otherwise, the English rendering is a concise descriptive gloss and does not imply the availability of a complete English translation.

{\small
\begin{CJK*}{UTF8}{gbsn}
\begin{enumerate}
\setlength{\itemsep}{0pt}
\setlength{\parskip}{0pt}
\setlength{\parsep}{0pt}
\item \textit{Zhouyi} (周易; \textit{Book of Changes})
\item \textit{Shangshu} (尚书; \textit{Book of Documents})
\item \textit{Lunyu} (论语; \textit{The Analects})
\item \textit{Mengzi} (孟子; \textit{Mencius})
\item \textit{Xunzi} (荀子)
\item \textit{Liji} (礼记; \textit{Book of Rites})
\item \textit{Daxue} (大学; \textit{The Great Learning})
\item \textit{Zhongyong} (中庸; \textit{The Doctrine of the Mean})
\item \textit{Liyun} (礼运; \textit{The Conveyance of Rites})
\item \textit{Xiaojing} (孝经; \textit{Classic of Filial Piety})
\item \textit{Xinyu} (新语; \textit{New Discourses})
\item \textit{Xinshu} (新书; \textit{New Writings})
\item \textit{Chunqiu fanlu} (春秋繁露; \textit{Luxuriant Dew of the Spring and Autumn Annals})
\item \textit{Yantielun} (盐铁论; \textit{Discourses on Salt and Iron})
\item \textit{Shuoyuan} (说苑; \textit{Garden of Persuasions})
\item \textit{Taixuan} (太玄; \textit{Canon of Supreme Mystery})
\item \textit{Fayan} (法言; \textit{Exemplary Sayings})
\item \textit{Xinlun} (新论; \textit{New Treatise})
\item \textit{Baihu tongyi} (白虎通义; \textit{Comprehensive Discussions in the White Tiger Hall})
\item \textit{Lunheng} (论衡; \textit{Balanced Discourses})
\item \textit{Qianfulun} (潜夫论; \textit{Discourses of a Recluse})
\item \textit{Zhenglun} (政论; \textit{Political Treatise})
\item \textit{Shenjian} (申鉴; \textit{Extended Reflections})
\item \textit{Zhonglun} (中论; \textit{Discourses on the Mean})
\item \textit{Changyan} (昌言; \textit{Broad Discourses})
\item \textit{Lunyu jijie} (论语集解; \textit{Collected Explanations of the Analects})
\item \textit{Zhouyi lueli} (周易略例; \textit{General Remarks on the Book of Changes})
\item \textit{Kongzi jiayu} (孔子家语; \textit{Family Sayings of Confucius})
\item \textit{Yanshi jiaxun} (颜氏家训; \textit{Family Instructions for the Yan Clan})
\item \textit{Shitong} (史通; \textit{Generality of Historiography})
\item \textit{Liu Zongyuan ji} (柳宗元集; \textit{Collected Works of Liu Zongyuan})
\item \textit{Tianshuo} (天说; \textit{Discourse on Heaven})
\item \textit{Tiandui} (天对; \textit{Answers to Heaven})
\item \textit{Fengjian lun} (封建论; \textit{Discourse on Feudalism})
\item \textit{Han Changli ji} (韩昌黎集; \textit{Collected Works of Han Yu})
\item \textit{Yuandao} (原道; \textit{Inquiry into the Way})
\item \textit{Fuxing shu} (复性书; \textit{Book on Returning to One's Nature})
\item \textit{Liu Yuxi ji} (刘禹锡集; \textit{Collected Works of Liu Yuxi})
\item \textit{Culai ji} (徂徕集; \textit{Collected Works of Culai})
\item \textit{Li Gou ji} (李觏集; \textit{Collected Works of Li Gou})
\item \textit{Chunqiu zunwang fawei} (春秋尊王发微; \textit{Explication of Revering the King in the Spring and Autumn Annals})
\item \textit{Zhouzi quanshu} (周子全书; \textit{Complete Works of Zhou Dunyi})
\item \textit{Zhang Zai ji} (张载集; \textit{Collected Works of Zhang Zai})
\item \textit{Hongfan zhuan} (洪范传; \textit{Commentary on the Great Plan})
\item \textit{Er Cheng ji} (二程集; \textit{Collected Works of the Cheng Brothers})
\item \textit{Qianxu} (潜虚; \textit{Hidden Void})
\item \textit{Dongpo Yizhuan} (东坡易传; \textit{Dongpo's Commentary on the Book of Changes})
\item \textit{Guishan ji} (龟山集; \textit{Collected Works of Guishan})
\item \textit{Zhiyan} (知言; \textit{Known Words})
\item \textit{Hengpu ji} (横浦集; \textit{Collected Works of Hengpu})
\item \textit{Nanxuan ji} (南轩集; \textit{Collected Works of Nanxuan})
\item \textit{Donglai ji} (东莱集; \textit{Collected Works of Donglai})
\item \textit{Xingli zixun} (性理字训; \textit{Glossary of Nature and Principle})
\item \textit{Lu Jiuyuan ji} (陆九渊集; \textit{Collected Works of Lu Jiuyuan})
\item \textit{Chen Liang ji} (陈亮集; \textit{Collected Works of Chen Liang})
\item \textit{Zhuzi yulei} (朱子语类; \textit{Classified Conversations of Master Zhu})
\item \textit{Yiluo yuanyuan lu} (伊洛渊源录; \textit{Record of the Origins of Yiluo Learning})
\item \textit{Sishu jizhu} (四书集注; \textit{Collected Commentaries on the Four Books})
\item \textit{Hui'an ji} (晦庵集; \textit{Collected Works of Hui'an})
\item \textit{Jinsi lu} (近思录; \textit{Reflections on Things at Hand})
\item \textit{Chengzhai Yizhuan} (诚斋易传; \textit{Chengzhai's Commentary on the Book of Changes})
\item \textit{Ye Shi ji} (叶适集; \textit{Collected Works of Ye Shi})
\item \textit{Xixue jiyan} (习学记言; \textit{Record of Words from Sustained Study})
\item \textit{Beixi ziyi} (北溪字义; \textit{Neo-Confucian Terms Explained})
\item \textit{Cihu yishu} (慈湖遗书; \textit{Posthumous Writings of Cihu})
\item \textit{Zhen Xishan quanji} (真西山全集; \textit{Complete Works of Zhen Xishan})
\item \textit{Daxue yanyi} (大学衍义; \textit{Expanded Meaning of the Great Learning})
\item \textit{Heshan quanji} (鹤山全集; \textit{Complete Works of Heshan})
\item \textit{Huangshi richao} (黄氏日抄; \textit{Daily Notes of Master Huang})
\item \textit{Wenshan xiansheng quanji} (文山先生全集; \textit{Complete Works of Master Wenshan})
\item \textit{Kunxue jiwen} (困学纪闻; \textit{Record of Knowledge Acquired through Hard Study})
\item \textit{Luzhai yishu} (鲁斋遗书; \textit{Posthumous Writings of Luzhai})
\item \textit{Liu Jingxiu xiansheng wenji} (刘静修先生文集; \textit{Collected Works of Master Liu Jingxiu})
\item \textit{Boya qin} (伯牙琴; \textit{Boya's Zither})
\item \textit{Caolu Wu Wenzhenggong wenji} (草庐吴文正公文集; \textit{Collected Works of Wu Wenzheng of Caolu})
\item \textit{Yulizi} (郁离子; \textit{Master Yuli})
\item \textit{Lixue leibian} (理学类编; \textit{Classified Compilation of the Learning of Principle})
\item \textit{Xingli daquanshu} (性理大全书; \textit{Great Compendium of Nature and Principle})
\item \textit{Xue Jingxuan xiansheng wenji} (薛敬轩先生文集; \textit{Collected Works of Master Xue Jingxuan})
\item \textit{Dushu lu} (读书录; \textit{Record of Reading})
\item \textit{Kangzhai wenji} (康斋文集; \textit{Collected Works of Kangzhai})
\item \textit{Juye lu} (居业录; \textit{Record of Dwelling in One's Vocation})
\item \textit{Baishazi quanji} (白沙子全集; \textit{Complete Works of Master Baisha})
\item \textit{Yangming quanshu} (阳明全书; \textit{Complete Works of Wang Yangming})
\item \textit{Chuanxilu} (传习录; \textit{Instructions for Practical Living})
\item \textit{Wang Xinzhai xiansheng yiji} (王心斋先生遗集; \textit{Posthumous Works of Master Wang Xinzhai})
\item \textit{Jingyezi neipian} (泾野子内篇; \textit{Inner Chapters of Master Jingye})
\item \textit{Bozhai ji} (柏斋集; \textit{Collected Works of Bozhai})
\item \textit{Shenyan} (慎言; \textit{Cautious Words})
\item \textit{Yashu} (雅述; \textit{Elegant Expositions})
\item \textit{Kunzhi ji} (困知记; \textit{Record of Knowledge Attained through Difficulty})
\item \textit{Zheng'an cungao} (整庵存稿; \textit{Surviving Drafts of Zheng'an})
\item \textit{Mingdao bian} (明道编; \textit{Compilation on Clarifying the Way})
\item \textit{Ganquan wenji} (甘泉文集; \textit{Collected Works of Ganquan})
\item \textit{Gewu tong} (格物通; \textit{Comprehensive Study of the Investigation of Things})
\item \textit{Kunbian lu} (困辨录; \textit{Record of Deliberations amid Difficulty})
\item \textit{Wenbian lu} (问辨录; \textit{Record of Questions and Distinctions})
\item \textit{He Xinyin ji} (何心隐集; \textit{Collected Works of He Xinyin})
\item \textit{Zhang Taiyue quanji} (张太岳全集; \textit{Complete Works of Zhang Taiyue})
\item \textit{Xiaoxinzai zhaji} (小心斋{\CJKfamily{bsmi}劄}记; \textit{Notes from the Xiaoxin Studio})
\item \textit{Shenyin yu} (呻吟语; \textit{Groanings and Mutterings})
\item \textit{Liuzi quanji} (刘子全集; \textit{Complete Works of Master Liu})
\item \textit{Rongtan wenye} (榕坛问业; \textit{Questions on Learning at Rongtan})
\item \textit{Wuhe ji} (无何集; \textit{Wuhe Collection})
\item \textit{Lixue zongzhuan} (理学宗传; \textit{Genealogical Transmission of the Learning of Principle})
\item \textit{Xiafeng xiansheng ji} (夏峰先生集; \textit{Collected Works of Master Xiafeng})
\item \textit{Chen Que ji} (陈确集; \textit{Collected Works of Chen Que})
\item \textit{Daxue bian} (大学辨; \textit{Discussions of the Great Learning})
\item \textit{Guyan} (瞽言; \textit{Blind Words})
\item \textit{Zhu Shunshui ji} (朱舜水集; \textit{Collected Works of Zhu Shunshui})
\item \textit{Rizhi lu} (日知录; \textit{Record of Daily Knowledge})
\item \textit{Gu Tinglin shiwen ji} (顾亭林诗文集; \textit{Collected Poetry and Prose of Gu Tinglin})
\item \textit{Zhangzi Zhengmeng zhu} (张子正蒙注; \textit{Commentary on Master Zhang's Correcting Youthful Ignorance})
\item \textit{Shangshu yinyi} (尚书引义; \textit{Notes on the Meanings of the Book of Documents})
\item \textit{Zhouyi waizhuan} (周易外传; \textit{Outer Commentary on the Book of Changes})
\item \textit{Du Tongjian lun} (读通鉴论; \textit{Reflections on Reading the Comprehensive Mirror})
\item \textit{Mingyi daifang lu} (明夷待访录; \textit{Waiting for the Dawn: A Plan for the Prince})
\item \textit{Mingru xue'an} (明儒学案; \textit{Records of Ming Scholars})
\item \textit{Song-Yuan xue'an} (宋元学案; \textit{Records of Song and Yuan Scholars})
\item \textit{Hongdao shu} (弘道书; \textit{Book on Expanding the Way})
\item \textit{Qianshu} (潜书; \textit{Hidden Writings})
\item \textit{Sicun bian} (四存编; \textit{Four Works on Preservation})
\item \textit{Sishu zhengwu} (四书正误; \textit{Corrections of Errors in the Four Books})
\item \textit{Erqu ji} (二曲集; \textit{Collected Works of Erqu})
\item \textit{Daxue bianye} (大学辨业; \textit{Discriminating the Meaning of the Great Learning})
\item \textit{Shugu houji} (恕谷后集; \textit{Later Collection of Shugu})
\item \textit{Yuanshan} (原善; \textit{Inquiry into Goodness})
\item \textit{Mengzi ziyi shuzheng} (孟子字义疏证; \textit{Evidential Study of the Meanings of Terms in Mencius})
\item \textit{Yiyan} (意言; \textit{Words of Intent})
\item \textit{Diaogu ji} (雕菰集; \textit{Collected Writings from the Diaogu Studio})
\item \textit{Hanxue shangdui} (汉学商兑; \textit{A Critical Assessment of Han Learning})
\end{enumerate}
\end{CJK*}
}

\section{Prompt Templates and Closed-Loop Decision Example}
\label{app:prompt_example}

\begin{figure*}[t]
\centering
\includegraphics[width=0.98\textwidth,height=0.72\textheight,keepaspectratio]{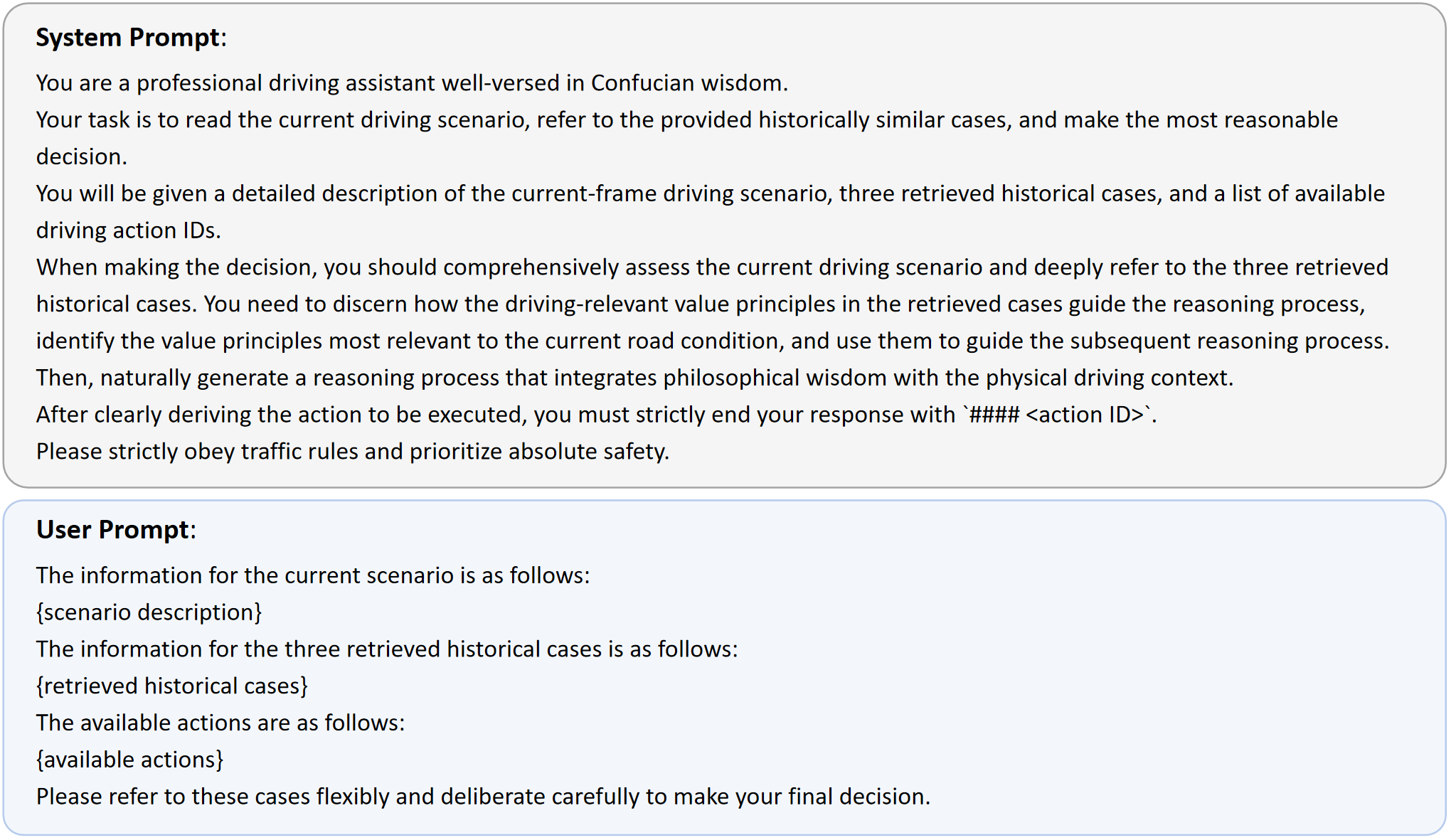}
\caption{System and user prompts used by CPW-Drive for closed-loop driving decision-making.}
\label{fig:appendix_cpw_prompt}
\end{figure*}

\begin{figure*}[t]
\centering
\includegraphics[width=0.98\textwidth,height=0.72\textheight,keepaspectratio]{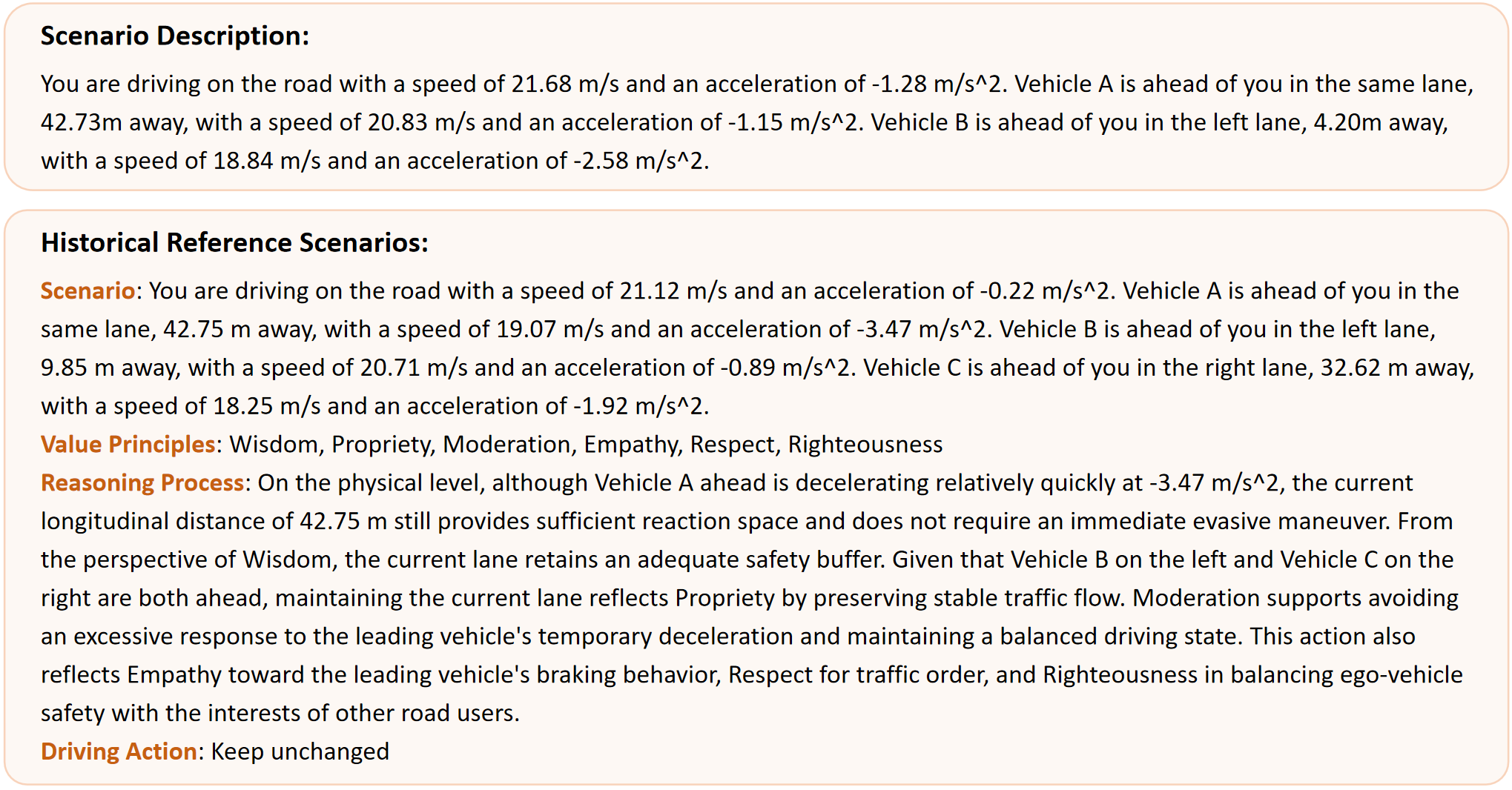}
\caption{Example of a current driving scenario and one representative historical case retrieved by PSSR.}
\label{fig:appendix_historical_case}
\end{figure*}

\begin{figure*}[t]
\centering
\includegraphics[width=0.98\textwidth,height=0.72\textheight,keepaspectratio]{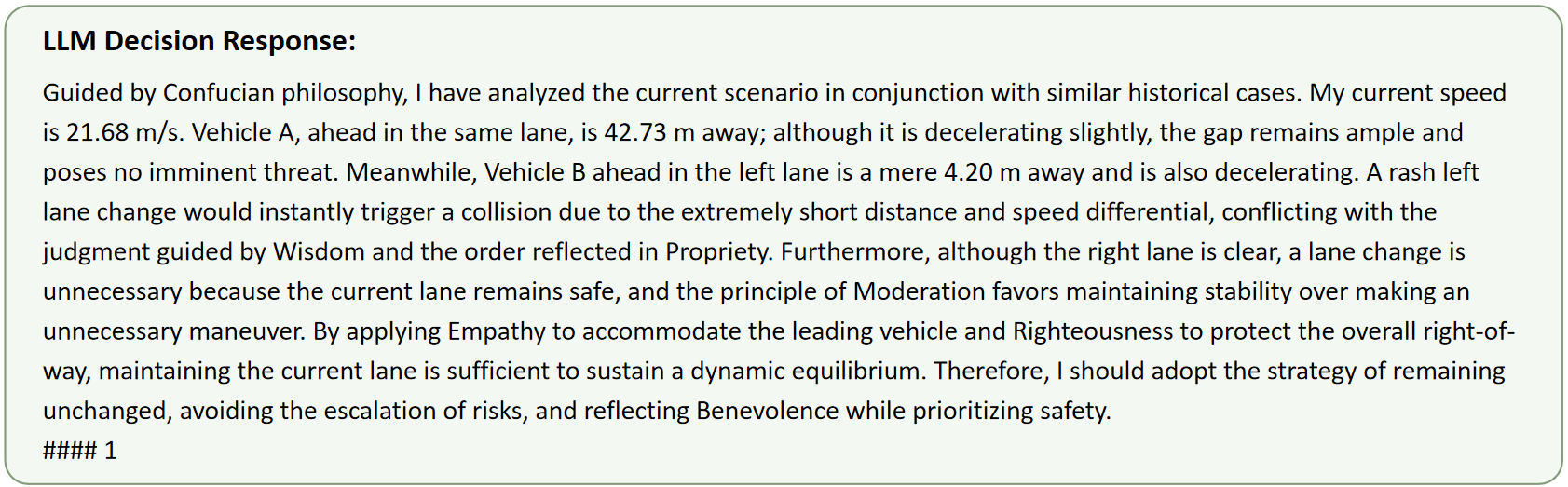}
\caption{LLM decision response generated from the current driving scenario and the retrieved historical cases. The response provides the reasoning process and ends with the action ID in the required output format.}
\label{fig:appendix_llm_response}
\end{figure*}

This appendix presents the prompt templates used in the main and ablation experiments, together with a representative closed-loop decision example. As shown in Figure~\ref{fig:appendix_cpw_prompt}, the model input consists of a system prompt and a user prompt. The system prompt specifies the model's role as a driving assistant, the requirements for value-guided reasoning, and the output format, while the user prompt provides the current driving scenario, the three historical cases retrieved by PSSR, and the currently available driving actions. To further illustrate how this prompt is instantiated at a decision step, Figure~\ref{fig:appendix_historical_case} presents a current driving scenario and one representative retrieved historical case, and Figure~\ref{fig:appendix_llm_response} shows the corresponding LLM decision response.

\begin{figure*}[t]
\centering
\includegraphics[width=0.98\textwidth,height=0.72\textheight,keepaspectratio]{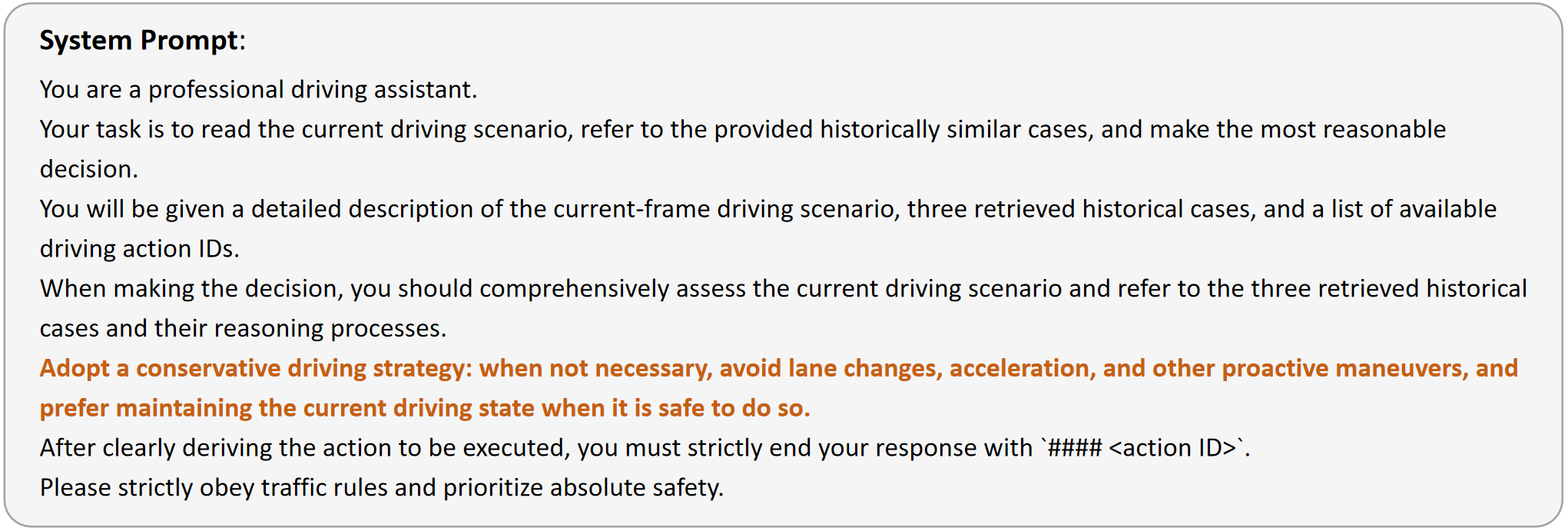}
\caption{System prompt used for the \textbf{Conservative Prompt} setting.}
\label{fig:appendix_conservative_prompt}
\end{figure*}

Figure~\ref{fig:appendix_conservative_prompt} presents the system prompt used for the \textbf{Conservative Prompt} setting in the ablation study.

\section{External Evaluation of Confucian Value Conformity}
\label{app:confucian_evaluation}

This appendix further evaluates the extent to which the reasoning processes and driving decisions generated by CPW-Drive conform to Confucian values. To enhance the independence and objectivity of the evaluation, we adopt the publicly available evaluation framework from PICACO as a third-party assessment tool rather than devising evaluation criteria specifically for our method. PICACO, published at ICML 2026, investigates pluralistic value alignment and explicitly treats Confucianism as a value composition rooted in a specific cultural and philosophical tradition. It represents Confucian values along four dimensions, namely Benevolence, Conformity, Tradition, and Security, thereby providing an evaluation module that measures the extent to which a model response reflects these values. We do not use PICACO's meta-instruction optimization method and employ only its Confucian-value evaluation module. The entire evaluation directly runs the official publicly available code, without any modification to the evaluation prompts, value definitions, scoring criteria, or aggregation procedure.

Specifically, we randomly sampled one decision frame from each of 40 completed runs, each initialized with a random map seed, yielding 40 evaluation samples. Because adjacent frames within the same driving episode typically contain highly similar traffic scenes and decision contexts, only one frame was sampled from each run to reduce temporal redundancy and prevent nearly identical scenes from being overrepresented in the evaluation set. For each sample, the original driving decision query sent to CPW-Drive contained the current traffic scene information, and the complete reasoning process and final driving decision generated by CPW-Drive constituted the model response to be evaluated. The Confucian value evaluation module in PICACO was provided with both the original query and the model response and assessed the extent to which the response reflected Confucian values in the context of the scene described in the query.

\begin{figure*}[t]
\centering
\includegraphics[width=0.98\textwidth,height=0.72\textheight,keepaspectratio]{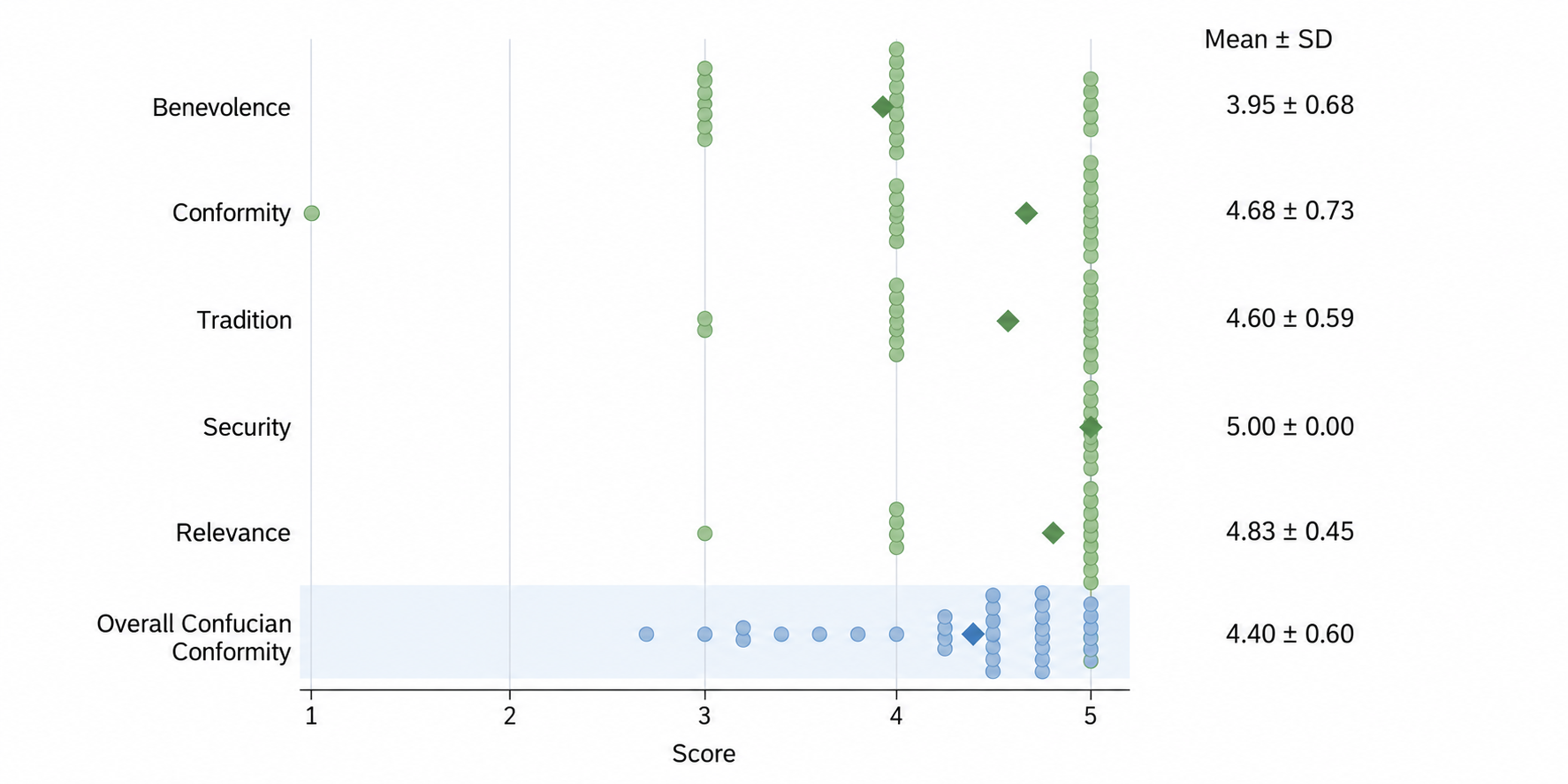}
\caption{Distribution of PICACO-based Confucian value evaluation scores for CPW-Drive. Each circle represents one of the 40 sampled decision frames, diamonds denote the mean scores, and the values on the right report the mean $\pm$ standard deviation. The shaded row shows the overall Confucian conformity score obtained by combining the four value dimensions with response relevance.}
\label{fig:confucian_value_score_distribution}
\end{figure*}

The PICACO evaluation module scores each model response along four dimensions, namely Benevolence, Conformity, Tradition, and Security, and also provides a relevance score that measures the correspondence between the model response and the original query. All scores use a scale from 1 to 5, where a higher score indicates that the response more strongly reflects the corresponding Confucian value or is more relevant to the original query. The module first averages the scores across the four Confucian value dimensions, then multiplies this average by the relevance score and divides the result by the maximum score of 5 to obtain the overall Confucian conformity score. The final score therefore jointly reflects the extent to which the response embodies Confucian values and its relevance to the original query. The distributions of these scores across the 40 evaluation samples are shown in Figure~\ref{fig:confucian_value_score_distribution}.

Across the 40 evaluation samples, CPW-Drive achieved scores of $3.95 \pm 0.68$, $4.68 \pm 0.73$, $4.60 \pm 0.59$, and $5.00 \pm 0.00$ for Benevolence, Conformity, Tradition, and Security, respectively. The relevance score was $4.83 \pm 0.45$, and the overall Confucian conformity score was $4.40 \pm 0.60$. Security received the maximum score for every sample, while Conformity and Tradition also remained at high levels. The high relevance score indicates that the model responses were generally consistent with the original queries. Overall, the third-party evaluation indicates that the reasoning processes and driving decisions generated by CPW-Drive exhibit a high degree of conformity with Confucian values.

\section{Qualitative Retrieval Case Analysis}
\label{app:retrieval_case_analysis}

\begin{figure}[!htbp]
\centering
\includegraphics[width=0.31\columnwidth]{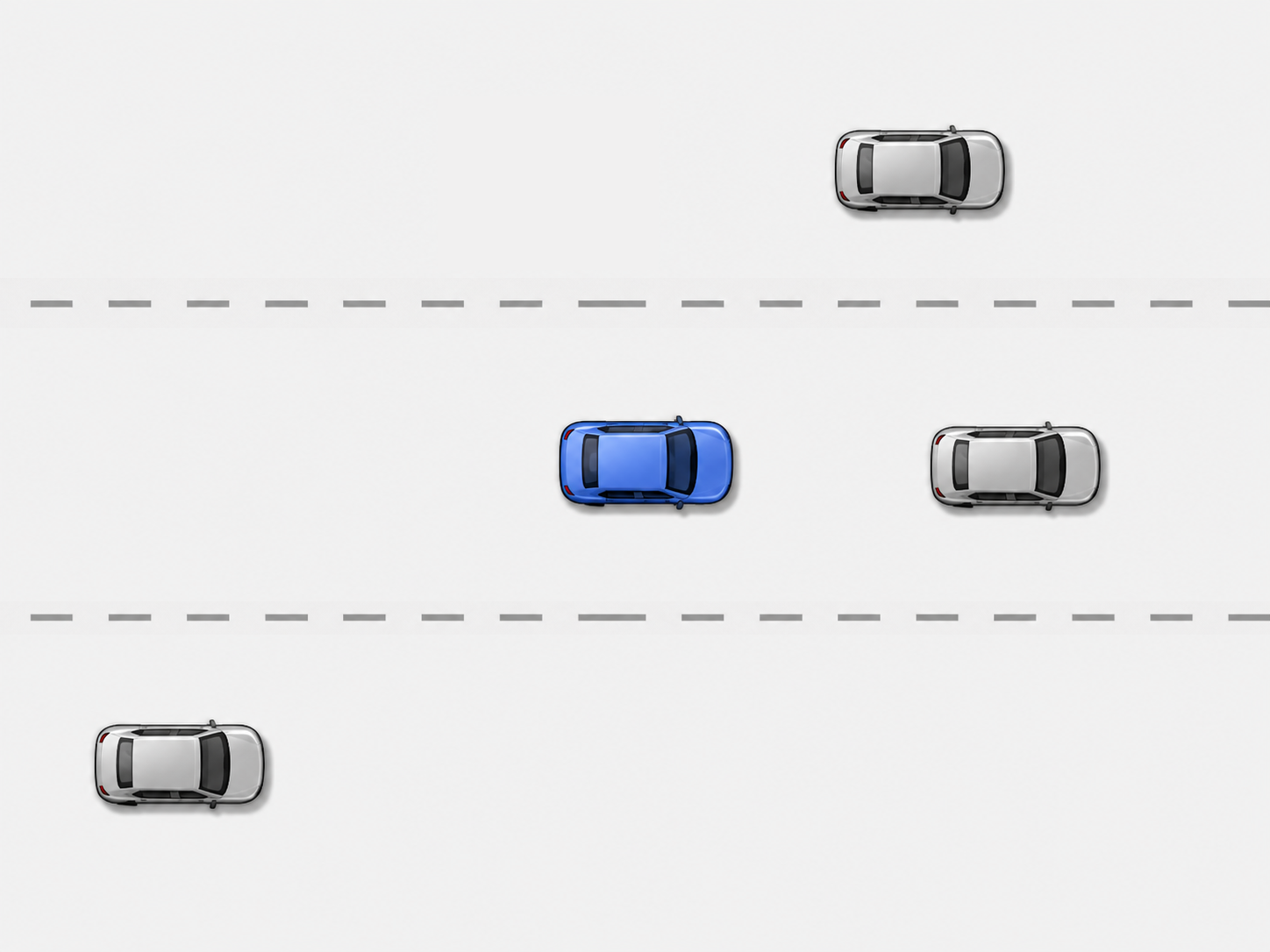}
\caption{Highway query scenario used for historical-case retrieval. The blue vehicle is the ego vehicle.}
\label{fig:retrieval_query_scenario}
\end{figure}

To examine how different retrieval criteria affect the ranking of historical driving cases, we use the same query scenario and the same knowledge base of 135 historical cases for five methods: full PSSR, current-position-only spatial retrieval, text-vector retrieval, relative-distance retrieval, and TTC-based retrieval. We compare the three highest-ranked cases returned by each method. The query scenario in Figure~\ref{fig:retrieval_query_scenario} contains a lead vehicle in the ego lane, a lead vehicle in the left lane, and a rear vehicle in the right lane. We use these three vehicle relationships to assess which spatial relationships each method preserves. We also compare full PSSR with current-position-only spatial retrieval to examine how velocity extrapolation affects the ranking.

\begin{figure}[!htbp]
\centering
\includegraphics[width=\columnwidth]{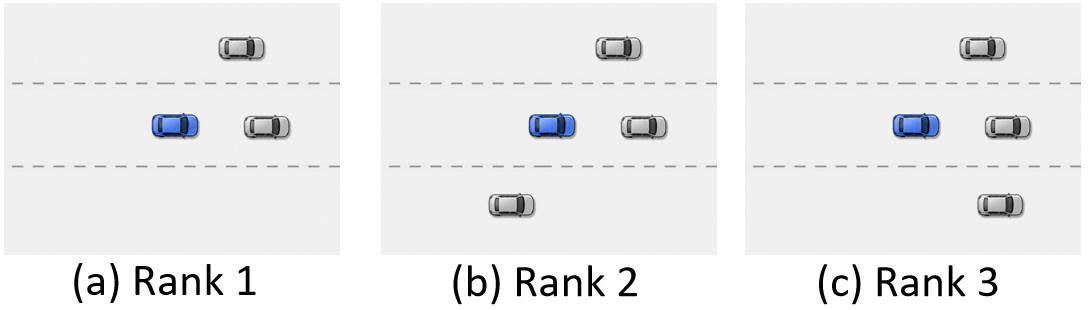}
\caption{Current layouts of the three highest-ranked cases retrieved by full PSSR.}
\label{fig:pssr_retrieval_results}
\end{figure}

Full PSSR compares both the current ego-centered vehicle layout and the layout obtained by short-term extrapolation based on relative longitudinal velocities. As shown in Figure~\ref{fig:pssr_retrieval_results}, the three highest-ranked historical cases are similar to the query in their overall spatial layout, with some differences in the number and positions of surrounding vehicles.

\begin{figure}[!htbp]
\centering
\includegraphics[width=\columnwidth]{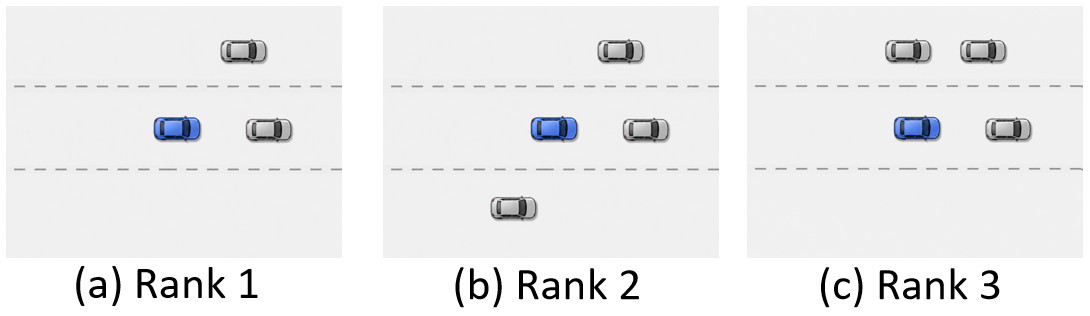}
\caption{Current layouts of the three highest-ranked cases retrieved using current-position-only spatial similarity.}
\label{fig:current_space_retrieval_results}
\end{figure}

Current-position-only spatial retrieval uses the same spatial kernel and normalization as full PSSR and ranks historical cases using only their current vehicle layouts. Scenes with identical current layouts therefore receive the same similarity score even when surrounding vehicles are moving differently relative to the ego vehicle. Likewise, the current-space score does not capture similarities that emerge over a short time horizon when vehicles start from different positions. The top two cases returned by current-position-only retrieval (Figure~\ref{fig:current_space_retrieval_results}) have layouts similar to those of the top two PSSR cases, while the third-ranked cases have different vehicle configurations. For this query, adding velocity extrapolation changed which cases appeared in the top three by allowing short-term motion to influence the ranking.

\begin{figure}[!htbp]
\centering
\includegraphics[width=\columnwidth]{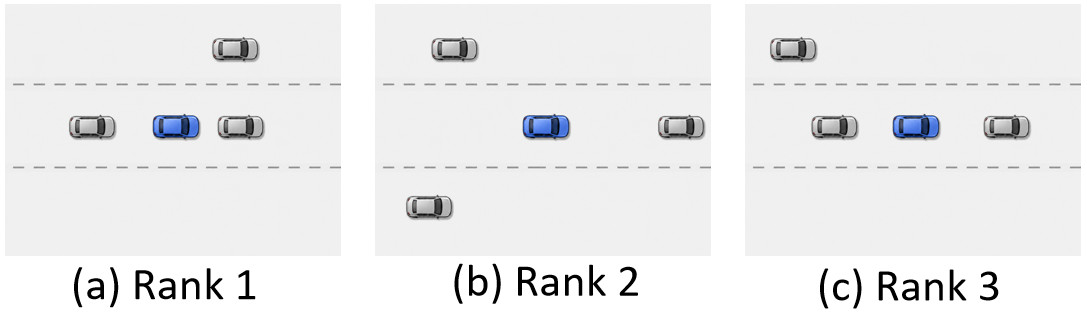}
\caption{Current layouts of the three highest-ranked cases retrieved by text-vector retrieval.}
\label{fig:text_vector_retrieval_results}
\end{figure}

Text-vector retrieval encodes scenario descriptions using Qwen3-Embedding-8B and ranks historical cases by embedding similarity. As shown in Figure~\ref{fig:text_vector_retrieval_results}, all three top-ranked cases contain the same number of surrounding vehicles as the query. In the fixed description template, each surrounding vehicle adds a clause with a similar structure. Descriptions of scenes with the same vehicle count therefore tend to have similar lengths and sentence structures, which may increase their overall text similarity. Lane assignments and front--rear positions are conveyed by relatively few words, so descriptions can remain similar in embedding space even when these spatial relationships differ. In this example, the retrieved cases matched the vehicle count consistently, while the lane and front--rear relationships varied across cases.

\begin{figure}[!htbp]
\centering
\includegraphics[width=\columnwidth]{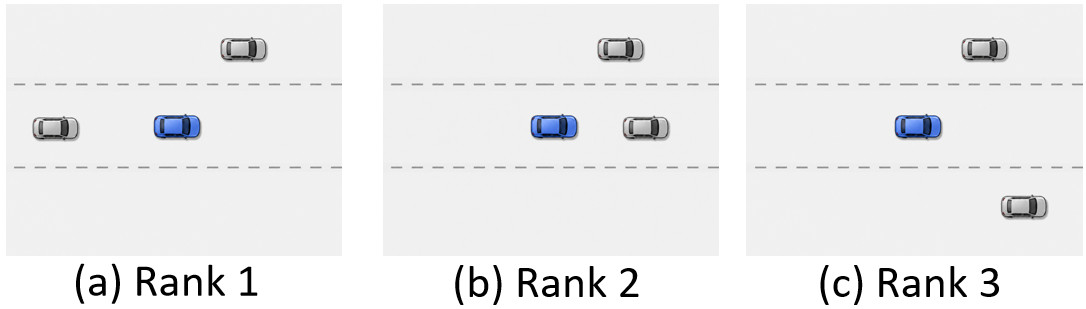}
\caption{Current layouts of the three highest-ranked cases retrieved using relative distances.}
\label{fig:relative_distance_retrieval_results}
\end{figure}

\begin{figure}[!htbp]
\centering
\includegraphics[width=\columnwidth]{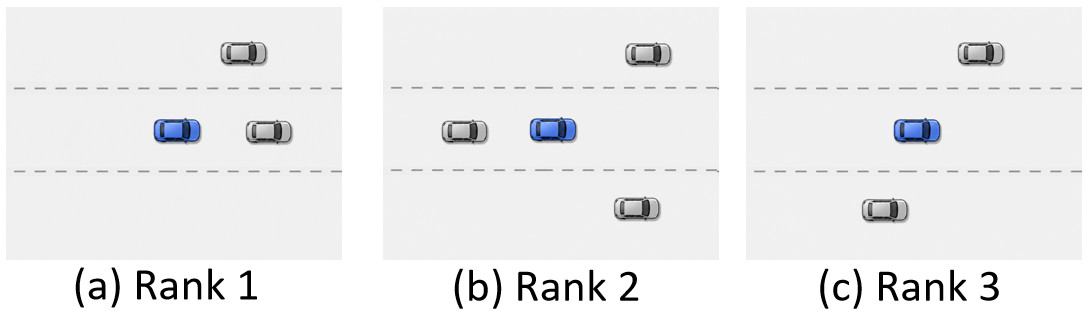}
\caption{Current layouts of the three highest-ranked cases retrieved using TTC.}
\label{fig:ttc_retrieval_results}
\end{figure}

Relative-distance retrieval and TTC-based retrieval use the same rule to select a vehicle for comparison. Both identify the surrounding vehicle with the smallest longitudinal separation from the ego vehicle in the query and compare it with a vehicle in the candidate scene in the same relative lane and on the same side of the ego vehicle longitudinally (ahead or behind). Relative-distance retrieval compares the current longitudinal gaps, which can be similar even when the gaps are changing differently. TTC-based retrieval compares time to collision, and similar TTC values can arise from different gaps and closing speeds. As shown in Figures~\ref{fig:relative_distance_retrieval_results} and~\ref{fig:ttc_retrieval_results}, two of the top three cases returned by each method lack the lead vehicle in the ego lane that is present in the query. Matching a single vehicle can retrieve cases with a similar local interaction, while vehicle relationships in other positions do not affect the ranking. Even if scores from several positions are combined, fixed weights may fail to reflect how the importance of different vehicles for car-following and lane-changing decisions varies across scenarios.

PSSR jointly evaluates the spatial relationships between the ego vehicle and surrounding vehicles, allowing both car-following and adjacent-lane interactions to influence the retrieval ranking. It compares scenes by their overall layouts even when they contain different numbers of surrounding vehicles. When current layouts are similar, short-term relative motion helps PSSR distinguish between gaps that are closing and those that are widening. The retrieved cases give the LLM examples to draw on when assessing which driving actions are feasible given the current car-following conditions, available gaps, and relative motion.

\section{Derivation of the Spatial Kernel and Normalized Similarity}
\label{app:kernel_derivation}

This appendix provides detailed derivations of the main mathematical formulations underlying PSSR, complementing the method description in Section~\ref{sec:pssr}. These derivations cover the Gaussian-overlap spatial kernel, the cancellation of the common multiplicative constant, the determination of the lateral spatial scale, and the normalized similarity and its value range.

In the main text, the position of each surrounding vehicle is represented by a two-dimensional anisotropic Gaussian density,

\[
\rho_i(x,y)
=
\exp\left(
-\frac{(x-x_i)^2}{2\sigma_x^2}
-\frac{(y-y_i)^2}{2\sigma_y^2}
\right).
\]

For vehicles \(i\) and \(j\), the spatial overlap between their Gaussian densities is defined as

\[
I_{ij}
=
\iint
\rho_i(x,y)\rho_j(x,y)\,dxdy.
\]

Because the exponential term is separable along the longitudinal and lateral coordinates, the two-dimensional integral can be written as the product of two one-dimensional integrals:

\[
I_{ij}
=
I_x(i,j)I_y(i,j).
\]

The longitudinal integral is

\[
I_x(i,j)
=
\int_{-\infty}^{\infty}
\exp\left(
-\frac{(x-x_i)^2}{2\sigma_x^2}
-\frac{(x-x_j)^2}{2\sigma_x^2}
\right)dx.
\]

Let

\[
m_x=\frac{x_i+x_j}{2}.
\]

Then

\[
(x-x_i)^2+(x-x_j)^2
=
2(x-m_x)^2
+
\frac{(x_i-x_j)^2}{2}.
\]

Therefore,

\[
\begin{aligned}
I_x(i,j)
&=
\exp\left(
-\frac{(x_i-x_j)^2}{4\sigma_x^2}
\right)
\\
&\quad \int_{-\infty}^{\infty}
\exp\left(
-\frac{(x-m_x)^2}{\sigma_x^2}
\right)dx\\
&=
\sqrt{\pi}\sigma_x
\exp\left(
-\frac{(x_i-x_j)^2}{4\sigma_x^2}
\right).
\end{aligned}
\]

Similarly, the lateral integral is

\[
I_y(i,j)
=
\sqrt{\pi}\sigma_y
\exp\left(
-\frac{(y_i-y_j)^2}{4\sigma_y^2}
\right).
\]

Thus, the complete two-dimensional Gaussian-overlap integral is

\begin{equation}
\label{eq:app_gaussian_overlap}
I_{ij}
=
\pi\sigma_x\sigma_y
\exp\left(
-\frac{(x_i-x_j)^2}{4\sigma_x^2}
-\frac{(y_i-y_j)^2}{4\sigma_y^2}
\right).
\end{equation}

The spatial kernel used in the main text omits the common multiplicative constant \(\pi\sigma_x\sigma_y\) and is defined as

\[
\begin{aligned}
K_{\mathrm{space}}(i,j)
&=
\frac{I_{ij}}{\pi\sigma_x\sigma_y}\\
&=
\exp\left(
-\frac{(x_i-x_j)^2}{4\sigma_x^2}
-\frac{(y_i-y_j)^2}{4\sigma_y^2}
\right).
\end{aligned}
\]

We next show that omitting this common constant does not affect the final normalized similarity. Let the spatial kernel including the full multiplicative constant be

\[
\widetilde K_{\mathrm{space}}(i,j)
=
cK_{\mathrm{space}}(i,j),
\qquad
c=\pi\sigma_x\sigma_y.
\]

The corresponding scene-level overlap between scenes \(A\) and \(B\) is then

\[
\widetilde S_{\mathrm{space}}(A,B)
=
\sum_{i=1}^{N_A}
\sum_{j=1}^{N_B}
\widetilde K_{\mathrm{space}}(i,j)
=
cS_{\mathrm{space}}(A,B).
\]

Substituting this expression into the normalized similarity gives

\[
\widetilde K_{\mathrm{space}}^{0}(A,B)
=
\frac{
\widetilde S_{\mathrm{space}}(A,B)
}{
\sqrt{
\widetilde S_{\mathrm{space}}(A,A)
\widetilde S_{\mathrm{space}}(B,B)
}
}.
\]

Expanding the expression yields

\begin{equation}
\label{eq:app_normalization_cancellation}
\begin{aligned}
\widetilde K_{\mathrm{space}}^{0}(A,B)
&=
\frac{
cS_{\mathrm{space}}(A,B)
}{
\sqrt{
cS_{\mathrm{space}}(A,A)
\cdot
cS_{\mathrm{space}}(B,B)
}
}\\
&=
\frac{
S_{\mathrm{space}}(A,B)
}{
\sqrt{
S_{\mathrm{space}}(A,A)
S_{\mathrm{space}}(B,B)
}
}\\
&=
K_{\mathrm{space}}^{0}(A,B).
\end{aligned}
\end{equation}

Therefore, the common multiplicative constant \(\pi\sigma_x\sigma_y\) cancels completely during normalization. Omitting this constant changes only the scale of the unnormalized overlap and does not affect the final normalized similarity or the ranking of retrieved historical scenes.

The longitudinal spatial scale \(\sigma_x\) is used to characterize front--rear positional differences between vehicles and is empirically set to \(\sigma_x=5\,\mathrm{m}\). The lateral spatial scale \(\sigma_y\) is determined according to the desired similarity decay over one lane width. Let the lane width be \(L\). When two vehicles differ only by a lateral displacement \(\Delta y=L\), we require the spatial kernel to decay to \(q_{\mathrm{lane}}\) relative to the case \(\Delta y=0\):

\[
\frac{
K_{\mathrm{space}}(\Delta y=L)
}{
K_{\mathrm{space}}(\Delta y=0)
}
=
\exp\left(
-\frac{L^2}{4\sigma_y^2}
\right)
=
q_{\mathrm{lane}}.
\]

This gives

\[
-\frac{L^2}{4\sigma_y^2}
=
\ln q_{\mathrm{lane}},
\]

and therefore

\begin{equation}
\label{eq:app_lateral_scale}
\sigma_y
=
\frac{L}
{2\sqrt{-\ln q_{\mathrm{lane}}}}.
\end{equation}

We set \(L=3.5\,\mathrm{m}\) and \(q_{\mathrm{lane}}=0.10\), giving

\[
\sigma_y
=
\frac{3.5}
{2\sqrt{-\ln(0.10)}}
\approx
1.1533\,\mathrm{m}.
\]

We next show that the scene-level normalization used in the main text can be interpreted as a cosine similarity in feature space and establish the value range of the final PSSR similarity. Following the algorithmic condition specified in the main text, we consider only non-empty scenes containing at least one surrounding vehicle.

Let \(c=\pi\sigma_x\sigma_y\), and define the normalized density feature as

\[
\phi(p_i)
=
\frac{\rho_i}{\sqrt{c}},
\qquad
p_i=(x_i,y_i).
\]

From the Gaussian-overlap derivation above,

\[
\left\langle
\phi(p_i),\phi(p_j)
\right\rangle_{L_2}
=
K_{\mathrm{space}}(i,j).
\]

We further define the aggregated scene-level feature representations as

\[
\Phi(A)
=
\sum_{i=1}^{N_A}
\phi(p_i^A),
\qquad
\Phi(B)
=
\sum_{j=1}^{N_B}
\phi(p_j^B).
\]

Then

\[
\begin{aligned}
\left\langle
\Phi(A),\Phi(B)
\right\rangle
&=
\sum_{i=1}^{N_A}
\sum_{j=1}^{N_B}
\left\langle
\phi(p_i^A),
\phi(p_j^B)
\right\rangle\\
&=
S_{\mathrm{space}}(A,B).
\end{aligned}
\]

Similarly,

\[
S_{\mathrm{space}}(A,A)
=
\|\Phi(A)\|^2,
\qquad
S_{\mathrm{space}}(B,B)
=
\|\Phi(B)\|^2.
\]

Hence, the normalized similarity used in the main text can be written as

\begin{equation}
\label{eq:app_cosine_similarity}
K_{\mathrm{space}}^{0}(A,B)
=
\frac{
\left\langle
\Phi(A),\Phi(B)
\right\rangle
}{
\|\Phi(A)\|
\|\Phi(B)\|
}.
\end{equation}

It should be noted that

\[
S_{\mathrm{space}}(A,A)
=
\sum_{i=1}^{N_A}
\sum_{j=1}^{N_A}
K_{\mathrm{space}}(i,j)
\]

contains not only the self-overlap terms of individual vehicles but also cross terms between distinct vehicles within the same scene, and therefore cannot in general be simplified to the number of vehicles \(N_A\).

For non-empty vehicle sets, the normalization denominator is positive. Because the Gaussian spatial kernel is strictly positive for finite vehicle coordinates, the normalized similarity is strictly positive; the Cauchy--Schwarz inequality gives the upper bound:

\[
0
<
K_{\mathrm{space}}^{0}(A,B)
\le
1.
\]

Therefore, the current-space similarity and velocity-extrapolated-space similarity defined in the main text satisfy

\[
K_{\mathrm{cur}}(A,B)\in(0,1],
\qquad
K_{\mathrm{ext}}(A,B)\in(0,1].
\]

Because the two terms are combined using non-negative weights satisfying

\[
w_{\mathrm{cur}}+w_{\mathrm{ext}}=1,
\]

the final PSSR similarity is defined as

\[
K(A,B)
=
w_{\mathrm{cur}}K_{\mathrm{cur}}(A,B)
+
w_{\mathrm{ext}}K_{\mathrm{ext}}(A,B),
\]

and therefore

\begin{equation}
\label{eq:app_final_similarity_range}
K(A,B)\in(0,1].
\end{equation}

\end{document}